\documentclass[twocolumn]{autart}    

\usepackage{graphicx}          

\usepackage{xcolor}

\usepackage{amsmath, amssymb,  amsfonts, enumitem}

\usepackage{bm}
\usepackage{color}
\usepackage{enumerate}
\usepackage{mathptmx}
\usepackage{bbm}
\usepackage{times}
\usepackage{lscape}   
\usepackage{booktabs} 
\usepackage{verbatim} 
\usepackage{mathtools}

\newcommand{\E}{\mathbb{E}}
\newcommand{\R}{\mathbb{R}}

\newcommand{\II}{\mathbb{I}}

\renewcommand{\paragraph}[1]{\vspace{0.15cm}\noindent {\bf #1}:}

\newcommand{\p}[1]{p\left(#1\right)}
\newcommand{\pc}[2]{\p{#1\vert #2}}

\newcommand{\lo}[1]{\log \left(#1\right)}
\newcommand{\NNo}[1]{\mathcal{N}\left(#1\right)}
\newcommand{\NN}[2]{\mathcal{N}\left(#1\vert #2\right)}

\newcommand{\EE}[1]{\E \left[#1\right]}
\newcommand{\EEE}[2]{\E_{#1} \left[#2\right]}

\newcommand{\LL}{\mathcal{L}}
\newcommand{\bs}[1]{\boldsymbol{#1}}

\newcommand{\parD}[2]{\frac{\partial #1}{\partial #2}}

\newcommand{\deT}[1]{\left\vert #1 \right\vert}
\newcommand{\KL}[2]{\text{KL}\left[#1\vert\vert #2\right]}
\newcommand{\sumall}[1]{\text{sum}\left[#1 \right]}

\newcommand{\sign}{\sigma_n^2}

\newcommand{\Kab}[2]{\bs{K}_{#1#2}}
\newcommand{\Koab}[2]{\overline{\bs{K}}_{#1#2}} 
\newcommand{\Qab}[2]{\bs{Q}_{#1#2}}

\newcommand{\Da}[1]{\bs{D}_{#1}}

\newcommand{\bR}{\bs{R}}
\newcommand{\bH}{\bs{H}}
\newcommand{\bX}{\bs{X}}
\newcommand{\bA}{\bs{A}}
\newcommand{\bB}{\bs{B}}
\newcommand{\bC}{\bs{C}}
\newcommand{\bD}{\bs{D}}
\newcommand{\bV}{\bs{V}}
\newcommand{\bQ}{\bs{Q}}
\newcommand{\bG}{\bs{G}}
\newcommand{\bS}{\bs{S}}

\newcommand{\bx}{\bs{x}}
\newcommand{\by}{\bs{y}}
\newcommand{\bff}{\bs{f}} 
\newcommand{\ba}{\bs{a}}
\newcommand{\bb}{\bs{b}}

\newcommand{\br}{\bs{r}}

\newcommand{\bu}{\bs{u}}
\newcommand{\bq}{\bs{q}}
\newcommand{\bv}{\bs{v}}

\newcommand{\bO}{\bs{0}} 

\newcommand{\bSig}{\bs{\Sigma}}
\newcommand{\bmu}{\bs{\mu}}
\newcommand{\bLam}{\bs{\Lambda}}
\newcommand{\bet}{\bs{\eta}}

\newcommand{\bthe}{\bs{\theta}}
\newcommand{\bThe}{\bs{\Theta}}
\newcommand{\bphi}{\bs{\phi}}

\newcommand{\DIAG}[1]{\text{Diag}\left[#1\right]}
\newcommand{\dIAG}[1]{\text{diag}\left[#1\right]}

\newcommand{\TR}[1]{\text{Tr}\left[#1\right]}
\newcommand*\diff{\mathop{}\!\mathrm{d}}

\def\app#1#2{%
	\mathrel{%
		\setbox0=\hbox{$#1\sim$}%
		\setbox2=\hbox{%
			\rlap{\hbox{$#1\propto$}}%
			\lower1.1\ht0\box0%
		}%
		\raise0.25\ht2\box2%
	}%
}
\def\approxprop{\mathpalette\app\relax}

\newtheorem{proposition}{Proposition}

\usepackage{hyperref}
\hypersetup{
	colorlinks=true,         
	urlcolor=blue,           
	pdfstartview=FitV
}
\begin{document}
	
	\begin{frontmatter}
		
		\title{Recursive Estimation for Sparse Gaussian Process Regression} 

		\author[M1,M2]{Manuel Sch\"urch}\ead{manuel@idsia.ch},    
		\author[M1]{Dario Azzimonti}\ead{dario.azzimonti@idsia.ch},               
		\author[M3,M1]{Alessio Benavoli}\ead{alessio.benavoli@ul.ie},
		\author[M1]{Marco Zaffalon}\ead{zaffalon@idsia.ch}  
		
		\address[M1]{Istituto Dalle Molle di Studi sull'Intelligenza Artificiale (IDSIA); Manno, Switzerland}  
		\address[M2]{Universit\`a della Svizzera italiana (USI); Lugano, Switzerland}
		\address[M3]{University of Limerick (UL); Limerick, Ireland}             


		\begin{abstract}                          
			
			Gaussian Processes (GPs) are powerful kernelized methods for non-parameteric regression used in many applications.
			However, their use is limited to a few thousand of training samples due to their cubic time complexity. 
			In order to scale GPs to larger datasets, several sparse approximations 
			based on so-called inducing points have been proposed in the literature.
			In this work we investigate the connection between a general class of sparse inducing point GP regression methods and Bayesian recursive estimation which enables Kalman Filter like updating for online learning. 
			The majority of previous work has focused on the batch setting, in particular for learning the model parameters and the position of the inducing points, 
			here instead we focus on training with mini-batches.
			By exploiting the Kalman filter formulation, we propose a novel approach that estimates such parameters by recursively propagating
			the analytical gradients of the posterior over mini-batches of the data. 
			Compared to state of the art methods, our method keeps analytic updates for the mean and covariance of the posterior, thus reducing drastically the size of the optimization problem.
			We show that our method achieves faster convergence and superior performance compared to state of the art sequential Gaussian Process regression on synthetic GP as well as real-world data with up to a million of data samples.
			
		\end{abstract}
		
	\end{frontmatter}
	
	\section{Introduction}

	\textit{Gaussian process} (GPs) regression
	is
	used in many applications, ranging from machine learning, social sciences, natural sciences and engineering, due to its modeling flexibility, robustness to overfitting and availability of well-calibrated predictive uncertainty estimates. 
	In control engineering, for example, GPs have been used in system identification for impulse response estimation \cite{pillonetto2010new,chen2012estimation,pillonetto2014kernel,pillonetto2016new}, nonlinear ARX models \cite{kocijan2005dynamic,Bijl2016OnlineSG},  learning of ODEs \cite{Barber2014GaussianPF,Macdonald2015},  latent force modeling \cite{Alvarez13} and to learn  the state space of a nonlinear dynamical system \cite{frigola2013bayesian,mattosrecurrent16,svensson2017flexible}. 
	However, GPs do not scale to large data sets due to their $\mathcal{O}(N^2)$ memory and $\mathcal{O}(N^3)$ computational costs, where $N$ is the number of training samples. For this reason several sparse GP approximations have been proposed in the literature. 
	Often such approximations are based on \textit{inducing points} methods, 
	where the unknown function is represented by its values at a set of $M \ll N$ pseudo-inputs, called inducing points.  
	Among such methods, \textit{Subset of Regressors} (SoR/DIC) approximations \cite{silverman1985some,wahba1999bias,smola2001sparse} produces overconfident predictions when leaving the training data.
	On the other hand, \textit{Deterministic Training Conditional} (DTC) \cite{csato2002sparse,seeger2003fast}, \textit{Fully Independent Training Conditional} (FITC) \cite{snelson2006sparse}, \textit{Fully Independent Conditional} (FIC)  \cite{quinonero2005unifying} and \textit{Partially  Independent  Training Conditional}  (PITC) \cite{quinonero2005unifying} all produce sensible uncertainty estimates. 
	These models differ from each other in the definition of their joint prior 
	over the latent function and test values. Titsias \cite{titsias2009variational}, instead, proposed to retain the exact prior but to perform approximate (variational) inference for the posterior, leading to the \textit{Variational Free Energy} (VFE) method which converges to full GP as $M$ increases.
	Bui et al. \cite{bui2016unifying} introduced \textit{Power Expectation Propagation} (PEP), based on the minimization of an $\alpha$-divergence, which unifies most of the previously mentioned models. 
	Typically, inference is achieved in $\mathcal{O}(M^2N)$ time and $\mathcal{O}(MN)$ space. 
	In order to find good parameters (inducing input points and kernel hyper-parameters), either the log marginal likelihood of the sparse models or a lower bound are numerically optimized. \\
	The previously mentioned approximations focus on the batch setting, i.e., all data is available at once and can be processed together. For big data, where the number of samples can be millions, keeping all data in memory is not possible, moreover the data might even arrive sequentially. 
	Bui et al. \cite{bui2017streaming} developed an algorithm to update hyper-parameters in an online fashion promising in a streaming setting, but with limited accuracy as each sample is considered only once.
	
	We focus here on the setting where hyper-parameters are learned by reconsidering mini-batches several times.
	In order to speed up the optimization, we would like to update the parameters more frequently for a subset of data and update the posterior in a sequential way. 
	In this setting, Hensman et al. \cite{hensman2013gaussian} 
	applied  \textit{Stochastic Variational Inference} (SVI,  \cite{hoffman2013stochastic}) to an \textit{uncollapsed} lower bound of the marginal likelihood. 
	The resulting  \textit{Stochastic Variational Gaussian Process} (SVGP) method allows to optimize the parameters with mini-batches. 
	Although showing high scalability and good accuracy, SVGP has two main drawbacks: 
	i) the (variational) posterior is not given analytically, which leads to 
	$ \mathcal{O}(M^2)$
	additional many parameters;
	ii) the uncollapsed  bounds are in practice often less tight than the corresponding collapsed VFE batch bounds because  
	the (variational) posterior 
	is not optimally eliminated. 
	The large number of parameters ($\approx MD+M^2$, where $D$ is the input space dimension) leads to a hard-to-tune optimization problem which requires appropriately decaying learning rates. Even for fixed parameters, each sample still needs to be reconsidered many times. 
	An orthogonal direction was pursued by the authors in \cite{hartikainen2010kalman} and \cite{sarkka2013spatiotemporal}, where a connection between GPs and State Space models for particular kernels was established for spatio-temporal regression problems, which allows to apply sequential algorithm such as the Kalman Filter,
	see also \cite{carron2016machine,benavoli2016b}.
	Inspired by this line of research, the authors in \cite{todescato2017efficient} focused on efficient implementation and extended the methodology to varying sampling locations over time. 
	These approaches can deal with sequential data and solve the problem of temporal time complexity, however the space complexity is still cubic in $N$. In addition, the hyper-parameters are usually fixed in advance.

	In this work we propose a \emph{recursive collapsed} lower bound to the log marginal likelihood which can be optimized stochastically with mini-batches. 
	
	In this respect, the first contribution of this paper is the derivation of a \textit{novel Kalman-filter-like} (KF)  formulation for a generic sparse inducing point method. In particular we show that sparse inducing point models can be seen as a Bayesian kernelized linear regression model with input dependent observation noise, a particular choice of basis functions and noise covariance. Given the model hyper-parameters, KF allows to train sparse GP methods analytically and exactly in an online setting (considering each sample only once, as opposed to the work in  \cite{hensman2013gaussian}  and \cite{hoang2015unifying}). In this formulation the posterior distribution obtained online is equivalent to full batch methods. This constitutes an interesting technique on its own for applications where  hyper-parameters are given, however the analysis above provides a key insight for parameter estimation.
	
	Our second main contribution is a \textit{recursive  approach
		to hyper-parameter estimation} based on the 
	KF formulation. It is based on recursively exploiting the chain rule for derivatives by recursively propagating the analytical gradients of the posterior which enables us to compute the derivatives of the lower bound sequentially. We show that, when computing the gradients of the recursive collapsed bound in a non-stochastic way, they exactly match the corresponding batch ones. 
	This new \textit{Stochastic Recursive Gradient Propagation} (SRGP)\footnote{Code is available at 
		\href{https://github.com/manuelIDSIA/SRGP}{https:\slash\slash github.com\slash manuelIDSIA\slash SRGP}.} approach constitutes an  efficient method to train 
	a very general class of sparse GP regression models
	with much fewer parameters to be estimated numerically ($\approx MD$)  than state of the art sequential GP regression methods ($\approx MD + M^2$). Since the number $M$ of inducing points determines the quality of the approximation to full GP, this reduction in number of parameters from $M^2$ to $M$ is crucial and   results in more accurate and faster convergence than state of the art approaches such as SVGP. For example, in the application to learn the input output behavior of a non-linear plan presented in Sect.~\ref{se:experiments} the number of parameters estimated by SVGP is $\approx 10500$ while our approach only estimates $\approx 500$ parameters due to the analytical updates.

	\section{Background on GP Regression}
	\label{se:GPR}
	Consider a training set 
	$\mathcal{D} = \left\{ y_i, \bx_i \right\}_{i=1}^N$ 
	of $N$ pairs of inputs $\bx_i\in \R^D$ and noisy scalar outputs $y_i$ generated by adding independent Gaussian noise to a latent function $f(\bx)$, that is $y_i = f(\bx_i)+\varepsilon_i$, where $\varepsilon_i\sim \NNo{0,\sign}$. 
	We denote by $\by = [y_1,\ldots,y_N]^T$ the vector of observations and by $\bX = [\bx_1^T,\ldots,\bx_N^T]^T \in \R^{N\times D}$ the input points.
	\\
	We model $f$ with a \textit{Gaussian Process} (GP), a stochastic process defined by its mean function $m(\bx)$ and covariance kernel $k(\bx,\bx')$.   
	The kernel $k$ is a positive definite function \cite{rasmussen2006gaussian}, 
	such as, for instance, 
	the \textit{squared exponential} (SE) kernel with individual lengthscales $l_i$ for each dimension, that is
	$k(\bx,\bx') 
	=\sigma_0^2 \exp\left(-\frac{1}{2}
	\left(\bx-\bx'\right)^T \DIAG{l_1^2,\ldots,l_D^2}
	\left(\bx-\bx'\right)
	\right)$. 
	We assume $m(x) \equiv 0$ for the sake of simplicity and we use the SE kernel throughout this paper however all methods work with any positive definite kernel. 
	Given the training 
	values $\bff = f\left(\bX\right)=\left[f(\bx_1),\ldots,f(\bx_N)\right]^T$
	and a test latent function value $f_*=f(\bx_*)$ at a test point $\bx_*\in \R^D$, then the joint distribution $p(\bff, f_*)$ is Gaussian. Our likelihood is Gaussian, $\pc{\by}{\bff} = \NN{\by}{\bff, \sign \II}$, and with Bayes theorem (see e.g. \cite{rasmussen2006gaussian}) we obtain analytically the posterior
	predictive 
	distribution
	$\pc{f_*}{\by} = \NN{f_*}{\bmu_*, \bSig_*}$ with
		\begin{align}
		\begin{split}
		\label{eq:predDist_full}
		\bmu_*&=
		\Kab{*}{\bX} \left( \Kab{\bX}{\bX} + \sign \II \right)^{-1} \by, 
		\\
		\bSig_*&=
		\Kab{*}{*} - \Kab{*}{\bX} \left( \Kab{\bX}{\bX} + \sign \II \right)^{-1} \Kab{\bX}{*},
		\end{split}
		\end{align}
	where 
	$\left[ \Kab{\bA}{\bB} \right]_{ij}=k(\ba_i,\bb_j)$ for any $\bA\in \R^{{M_1}\times D}$ and $\bB\in \R^{{M_2}\times D}$ with the corresponding rows $\ba_i,\bb_j$. For brevity, we use $*$ to indicate $\bx_*$. 
	The GP depends via the kernel matrices on the 
	hyper-parameters $\bphi = \{\sigma_0,l_1,\ldots,l_D,\sigma_n\}$ typically estimated by maximizing the log marginal likelihood
	\begin{align}
	\label{eq:GP_lik}
	\log \pc{\by}{\bphi} 
	= 
	\log \NN{\by}{\bO, \Kab{\bX}{\bX} + \sign \II}. 
	\end{align}
	Note that the computations for inference require the inversion of the  matrix in Eq. \eqref{eq:predDist_full} 
	which scales as 
	$\mathcal{O}(N^3)$ in time and $\mathcal{O}(N^2)$ for memory (given $\bphi$). 
	
	\subsection{Batch Sparse GP Regression}
	\label{sse:sparse_GP}
\begin{figure}[htb!]
		\begin{center}
			\includegraphics[width=0.48\textwidth]{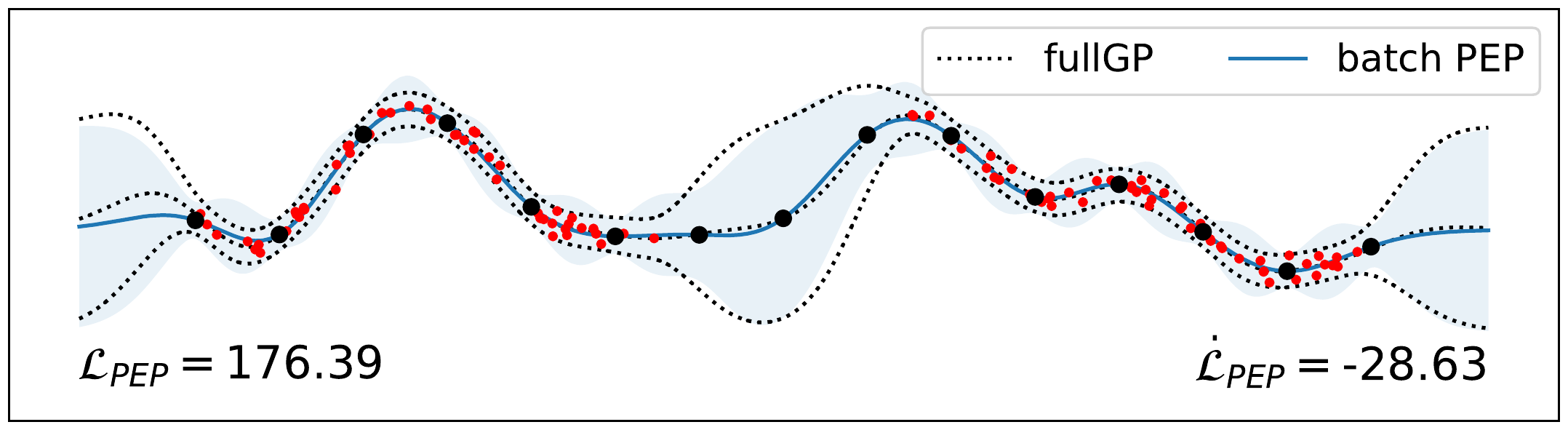}
			\caption{Full GP and batch sparse GP regression with PEP model ($\alpha=0.5$). $N=100$ data samples are summarized with $15$ equidistant inducing points (black dots). A slightly smaller than optimal lengthscale was selected and no parameters $\bthe$ were optimized. 
				The numbers in the left and right corner indicate the lower bound to the log marginal likelihood in \eqref{eq:PEP_log} and its derivative with respect to the lengthscale, respectively.
			}
			\label{fig_sparsevsfull}
		\end{center}
	\end{figure}

	Sparse GP regression methods based on \textit{inducing points} approaches reduce the computational complexity by introducing  
	$M \ll N$ inducing points $\bu \in \R^M$ 
	that optimally summarize the dependency of the whole training data.  The inducing \textit{inputs} $\bR \in \R^{M\times D}$ are in the $D$-dimensional input data space and the inducing \textit{outputs} $\bu := f(\bR)$ are the corresponding GP-function values, see also Fig. \ref{fig_sparsevsfull}.
	The GP prior over $\bff$ and $f_*$ is augmented 
	with the 
	inducing outputs $\bu$, leading to a joint $\p{\bff,f_*,\bu}$
	and marginal $\p{\bu} = \NNo{\bO,\Kab{\bR}{\bR}}$ prior. 
	By marginalizing out the inducing points, the original prior 
	$\p{\bff,f_*} =
	\int 
	\pc{\bff,f_*}{\bu} \p{\bu} \diff\bu
	$
	is recovered.
	The fundamental approximation in 
	all sparse GP models is that given the inducing outputs $\bu$, $\bff$ and $f$ are conditionally independent. 
	Consequently, inference in these models can be done in $\mathcal{O}(M^2N)$ time and $\mathcal{O}(MN)$ space \cite{snelson2006sparse}. 

	We briefly recall here the  sparse predictive distribution and the variational lower bound to the log marginal likelihood for the \textit{Power Expectation Propagation} (PEP) model \cite{bui2016unifying} because it unifies the main sparse inducing points approaches. 
	The variational lower bound is used for optimizing the parameters $\bthe := \{\bphi, \bR\}$. 
	In the following, we denote
	$\Qab{\bA}{\bB} = \Kab{\bA}{\bR}\Kab{\bR}{\bR}^{-1}\Kab{\bR}{\bB}$
	and 
	$\Da{\bA}=\Kab{\bA}{\bA} - \Qab{\bA}{\bA}$
	for  any 
	$\bA,\bB$. 
	The predictive distribution $\pc{f_*}{\by} = \NN{f_*}{\bmu_*, \bSig_*}$ of PEP
	is given by
	\begin{align}
	\label{eq:PEP_pred}
	\begin{split}
	\bmu_*&=
	\Qab{*}{\bX} \left( \Koab{\bX}{\bX} + \sign \II \right)^{-1} \by,\\
	\bSig_*&=
	\Kab{*}{*} - \Qab{*}{\bX} \left( \Koab{\bX}{\bX} + \sign \II \right)^{-1} \Qab{\bX}{*},
	\end{split}
	\end{align}
	where 
	$
	\Koab{\bX}{\bX}
	=\Qab{\bX}{\bX}
	+
	\alpha \DIAG{\Da{\bX}}$. 
	A lower bound to the sparse log marginal likelihood is analytically available
	\begin{align}
	\label{eq:PEP_log}
	\begin{split}
	\LL_{PEP}(\bthe)
	=& 
	\log \NN{\by}{\bO, \Koab{\bX}{\bX} + \sign \II} \\
	&-\frac{1-\alpha}{2\alpha} 
	\sum_{i=1}^N
	\lo{ 1+ \frac{\alpha}{\sign} \left[ \Da{\bX}\right]_{ii} },
	\end{split}
	\end{align}
	where we omit the explicit dependency on $\bthe$ via $ \Koab{\bX}{\bX}$ and $\Da{\bX}$ for the sake of brevity. This bound can be used to learn the parameters $\bthe$, similarly to Eq.~\eqref{eq:GP_lik} for full GP. 

	The special case $\alpha\rightarrow 0$ was  originally introduced in \cite{titsias2009variational} where the author proposed to maximize a variational lower bound to the true GP marginal likelihood, obtaining
	the \textit{Variational Free Energy} (VFE) or the \textit{collapsed lower bound} 
	\begin{align}
	\label{eq:collapsed_bound}
	\LL_{VFE}(\bthe)
	= 
	\log \NN{\by}{\bO, \Qab{\bX}{\bX} + \sign \II}
	-\frac{\TR{\Da{\bX}}}{2\sign}.
	\end{align}
	In \eqref{eq:collapsed_bound} the variational distribution 
	over the inducing points is optimally eliminated and analytically available. 
	The rightmost term in \eqref{eq:collapsed_bound}
	acts as a regularizer that prevents overfitting and has the effect that the sparse GP predictive distribution  \eqref{eq:PEP_pred}
	converges \cite{titsias2009variational} to the exact GP predictive distribution \eqref{eq:predDist_full} as the number of inducing points increases, when optimizing $\bthe$ with \eqref{eq:collapsed_bound}. 
	See also \cite{bui2016unifying,liu2018gaussian,quinonero2005unifying,rasmussen2006gaussian} for recent reviews on the subject. 
	
	\subsection{Sequential Sparse GP Regression}
	\label{sse:SVI}
	The optimization for $\bthe$ of the collapsed lower bound \eqref{eq:collapsed_bound} requires to process the whole dataset, which is very inefficient and not feasible for large $N$. We would like to update the parameters more frequently, therefore,
	we split the data $\mathcal{D} = \left\{ \by_k, \bX_k \right\}_{k=1}^K$  into $K$ mini-batches of size $B$ and denote $\bff_k$ the corresponding sparse GP value. 
	\textit{Stochastic Variational Gaussian Process} (SVGP) \cite{hensman2013gaussian} achieves this result by
	applying stochastic optimization
	to an \textit{uncollapsed lower bound}  to the log marginal likelihood
	\begin{align}
	\begin{split}
	\label{eq:uncollapsed_bound}
	&\LL_{SVGP}(\bmu,\bSig,\bthe)
	=  -\KL{\bq(\bu)}{\pc{\bu}{\bthe}}\\
	&+K \sum_{k=1}^{K} \int 
	\bq(\bu) \pc{\bff_k}{\bu,\bthe} \pc{\by_k}{\bff_k,\bthe} \diff \bu,
	\end{split}
	\end{align}
	where the variational distribution $q(\bu)$ is part of the bound and explicitly parametrized as $\bq(\bu)=\NN{\bu}{\bmu,\bSig}$. 
	This uncollapsed bound satisfies $\LL_{SVGP}(\bmu,\bSig,\bthe)\leq \LL_{VFE}(\bthe)$ with equality when inserting the optimal mean and covariance of the variational distribution of VFE. The key property of this bound is that it can be written as a sum of $K$ terms, which allows \textit{Stochastic Variational Inference} (SVI, \cite{hoffman2013stochastic}). Note that collapsing the bound, i.e. inserting the optimal distribution, reintroduces dependencies between the observations, and eliminates the global parameter $\bu$ which is needed for SVI. For this reason,  all variational parameters are numerically estimated by following the noisy gradients of a stochastic estimate of the lower bound $\LL_{SVGP}$. 
	By passing through the training data a sufficient number of times, the variational distribution 
	converges to the batch solution of VFE method.
	This approach, however, requires a large number of parameters: in addition to the parameters $\bthe$, all entries in the mean vector $\bmu$ and the covariance matrix $\bSig$ have to be estimated numerically, which is in order $\mathcal{O}(M^2)$.

	\section{Recursive Sparse GP Regression}
	\label{se:bayes_est}
	In this section we establish the connection between Bayesian recursive estimation and sparse inducing point GP models. 
	We recall the \textit{weight-space view}  for a large class of sparse inducing point GP models, which we present here as a particular kernelized version of a Bayesian linear regression model. See \cite[Ch. 2.1]{rasmussen2006gaussian},  for an analogous discussion on the full GP model. 
	Here, however, we show how to exploit the  KF 
	to train many sparse methods analytically either in an online setting for fixed hyper-parameters. This allows us to introduce a recursive log marginal likelihood with a model specific regularization term for parameter estimation.

	\subsection{Weight-Space View of Generic Sparse GP}
	\label{se:weight_view}
\begin{figure}[htp!]
		\centering
		\includegraphics[width=0.48\textwidth]{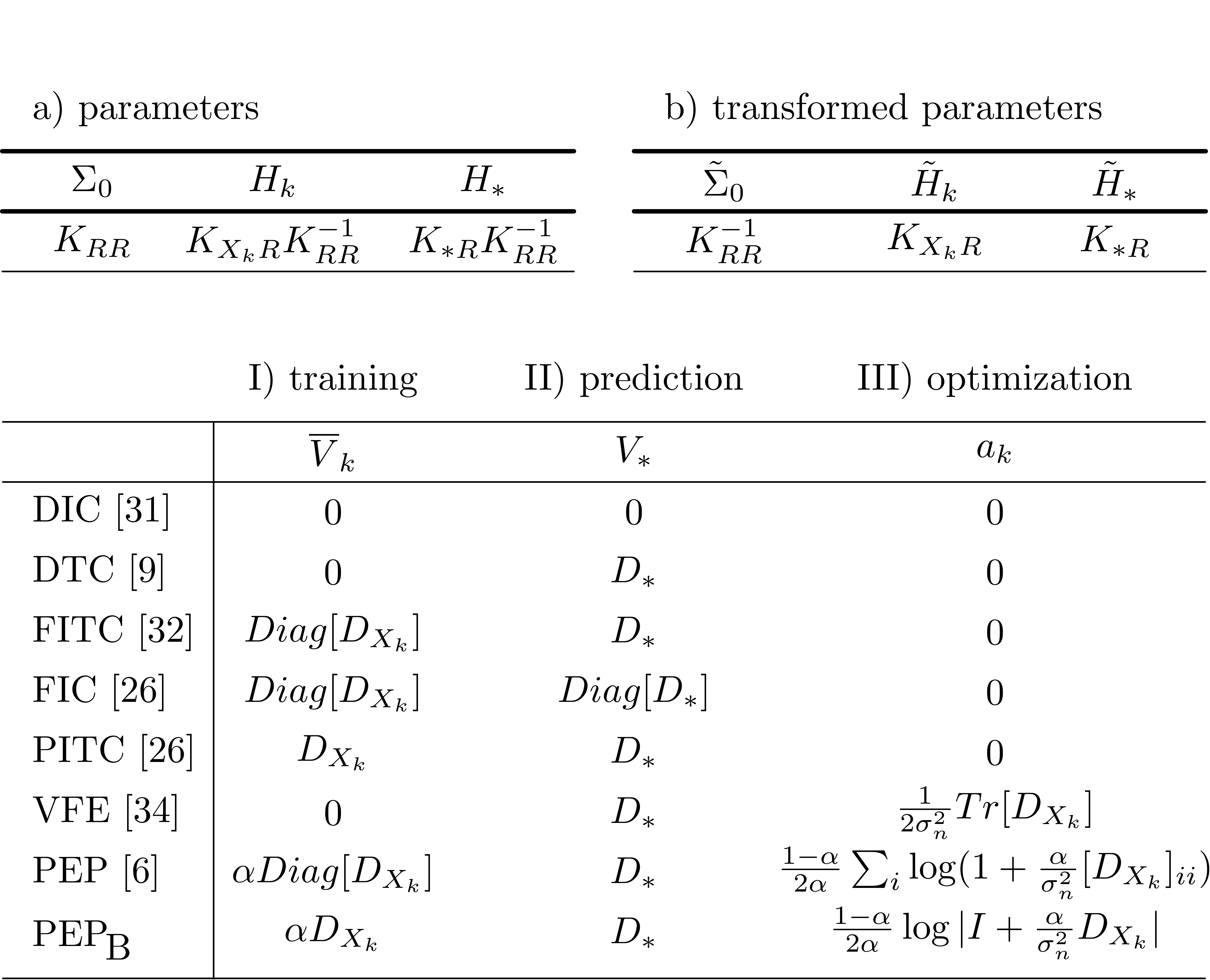}
		\caption{Summary of parameters for sparse GP models for recursive estimation. For all models, we have
			$\bmu_{0}=\bO,$
			$\bV_k = \overline{\bV}_k + \sign \II $ and
			$\bSig_{0}$, $\bH_k$,
			$\bH_{*}$ from the table a) or a transformed version b). Using the model specific quantities for the observation noise $\overline{\bV}_k$, the prediction covariance $\bV_*$ and the regularization term $a_k$ from the bottom table allows the training with the recursive approaches.
		}
		\label{fig:parameters}
	\end{figure}

	For a mini-batch $\bX \in \R^{B\times D}$ of size $B$, consider the generic sparse GP model
	\begin{align}
	f\left(\bX\right) &= H\left(\bX\right) \bu + \gamma\left(\bX\right)
	\end{align}
	where the sparse GP value 
	$f\left(\bX\right)$  is modeled by a linear combination of basis-functions 
	$H\left(\bX\right) \in \R^{B\times M}$, (stochastic) weights $\bu\in \R^{M}$ 
	with a prior $\p{\bu} = \NNo{\bO, \bSig_0}$ 
	and  an input dependent error term $\ \gamma\left(\bX\right) \sim \NNo{0,V\left(\bX\right)}$ that takes into account the sparse approximation. 
	For $k=1,\ldots,K$, the noisy observations $\by_k$ are obtained by adding independent noise $\bs{ \varepsilon}_k \sim \NNo{0, \sign \II}$ to
	$f\left(\bX_k\right)$, yielding the model
	\begin{align}
	\by_k &= \bff_k + \bs{\varepsilon}_k ;
	\\
	\bff_k = \bH_k \bu + \bs{\gamma}_k
	~~&\text{ and }~~
	\bff_* = \bH_* \bu + \bs{\gamma}_*,
	\end{align}
	where we distinguish the training $\bff_k=f\left(\bX_k\right)$ and test $\bff_*=f\left(\bX_*\right)$ cases 
	depending on the input $\bX_k$ and $\bX_*$.
	Assuming 
	$\bs{ \gamma}_k $, $\bs{ \gamma}_* $ and $\bs{ \varepsilon}_k$  are independent,
	by linearity and Gaussianity we can compactly write
	\begin{align}
	\label{eq:pyf}
	\pc{\by_k}{\bff_k} &= \NN{\by_k}{\bff_k, \sigma_n^2 \II};
	\\
	\label{eq:pu}
	\p{\bu} &= \NNo{\bO, \bSig_0};
	\\
	\label{eq:fu}
	\pc{\bff_k}{\bu}&=\NN{\bff_k}{\bH_k\bu,\overline{\bV}_k};
	\\
	\pc{\bff_*}{\bu}&=\NN{\bff_*}{\bH_*\bu,\bV_*},
	\end{align}
	Combining \eqref{eq:pyf} and \eqref{eq:fu} and
	by integrating out $\bff_k$ 
	we obtain the likelihood 
	$
	\pc{\by_k}{\bu} = \NN{\by_k}{ \bH_k \bu, \bV_k}
	$ 
	where $\bV_k=\overline{\bV}_k+\sigma_n^2 \II$. This shows that a generic sparse GP regression model
	can be seen as a Bayesian non-linear regression model with additional input dependent observation noise, a particular choice of basis functions 
	$\bH_k$ 
	and covariance structures $\overline{\bV}_k$, $\bV_*$.
	For inducing inputs $\bR\in \R^{M\times D}$,
	we have
	$\bSig_0 = \Kab{\bR}{\bR}$,
	$\bH_k = \Kab{\bX_k}{\bR}  \Kab{\bR}{\bR}^{-1}$
	and
	$\bH_* = \Kab{*}{\bR}  \Kab{\bR}{\bR}^{-1}$. 
	Different choices of the quantities $\overline{\bV}_k$ and $\bV_*$ lead to a range of sparse GP models summarized in the bottom table in Fig.~\ref{fig:parameters}. 

	\subsection{Training}
	Given the prior $\p{\bu}$ and the likelihood $\pc{\by_k}{\bu}$, the posterior over the weights $\bu$ conditioned on the data $\by_{1:k}$ can be  computed either in a batch or in a recursive manner.
	
	\subsubsection{Batch Estimation}
	
	The batch likelihood is
	$\pc{\by}{\bu}=
	\prod_{k=1}^K \pc{\by_k}{\bu}
	=\NN{\by}{\bH\bu,\bV}$ 
	with 
	$
	\bH = \left[
	\bH_1^T,\ldots,\bH_K^T
	\right]^T \in \R^{N\times M}$
	and
	$ \bV $ a block-diagonal matrix with blocks $\bV_k$. 
	The posterior over $\bu$ 
	given the data $\by$ 
	can be obtained by Bayes' rule, i.e. 
	\begin{align}
	\label{eq:batch}
	\pc{\bu}{\by} 
	\propto
	\pc{\by}{\bu} \p{\bu} 
	&\propto \NN{\bu}{\bmu_K, \bSig_K},
	\end{align}
	with
	$\bSig_K
	= \left( \bSig_0^{-1} +
	\bH^T \bV^{-1} \bH
	\right)^{-1}
	$
	and
	$
	\bmu_K=
	\bSig_K \bH^T \bV^{-1} \by.
	$

	\subsubsection{Recursive Estimation}
	\label{sse:rec_est}
	An equivalent solution can be obtained by propagating recursively
	$\pc{\bu}{\by_{1:k-1}}$. By interpreting this previous posterior as the prior, the updated posterior can be recursively computed by 
	\begin{align}
	\begin{split}
	\label{eq:rec}
	\pc{\bu}{\by_{1:k}} 
	\propto
	\pc{\by_k}{\bu} \pc{\bu}{\by_{1:k-1}}
	\propto \NN{\bu}{\bmu_k, \bSig_k}, 
	\end{split}
	\end{align}
	where
	$
	\bmu_k=
	\bSig_k
	\left(
	\bH_k^T \bV_k^{-1} \by_k
	+
	\bSig_{k-1}^{-1}\bmu_{k-1}
	\right)
	$
	and
	$
	\bSig_k
	= \left( \bSig_{k-1}^{-1} +
	\bH_k^T \bV_k^{-1} \bH_k
	\right)^{-1}
	$.
	%
	
	\paragraph{Kalman Filter like updating}
	the KF constitutes an efficient way to update the mean and covariance of $\pc{\bu}{\by_{1:k}} $.
	Applying 
	the Woodbury identity 
	to $\bSig_k$ in Eq. \eqref{eq:rec} 
	and introducing temporary variables 
	yields 
	\begin{align}
	\label{eq:KF}
	\begin{split}
	\br_k &= \by_k - \bH_k \bmu_{k-1}; \\
	\bS_k& = \bH_k  \bSig_{k-1} \bH_k^T + \bV_k;\\
	\bG_k &= \bSig_{k-1} \bH_k^T  \bS_k^{-1};
	\end{split}
	\begin{split}
	\bmu_{k} &= \bmu_{k-1} + \bG_k\br_k;\\
	\bSig_{k} &= \bSig_{k-1} - \bG_k \bS_k \bG_k^T.
	\end{split}
	\end{align}
	Starting the recursion with
	$\bmu_0=\bO$
	and
	$\bSig_0$, 
	the posterior distribution at step $K$ is equivalent to \eqref{eq:batch} independent of the order of the data. 
	We want to emphasize that the only difference in the estimation part between the sparse GP models is the form of the additional noise $\bV_k= \overline{\bV}_k + \sign \II$.
	%
	%
	
	\paragraph{Transformation}
	\label{sse:trans0}
	instead of running a KF with 
	$\bSig_0 = \Kab{\bR}{\bR}$,
	$\bH_k = \Kab{\bX_k}{\bR}  \Kab{\bR}{\bR}^{-1}$
	and
	$\bH_* = \Kab{*}{\bR}  \Kab{\bR}{\bR}^{-1}$, an equivalent predictive distribution is also obtained when using
	$\tilde{\bSig}_0 = \Kab{\bR}{\bR}^{-1}$ and $\tilde{\bH}_k =\Kab{\bX_k}{\bR}$ together with 
	$\tilde{\bH}_* = \Kab{*}{\bR} $. 
	For any $k$ we then propagate a transformed posterior distribution 
	$ \tilde{\bmu}_k = \Kab{\bR}{\bR}^{-1}\bmu_k $,
	$\tilde{\bSig}_k =  \Kab{\bR}{\bR}^{-1} \bSig_k \Kab{\bR}{\bR}^{-1}  $
	and
	$ \bmu_k= \Kab{\bR}{\bR} \tilde{\bmu}_k$,
	$ \bSig_k=  \Kab{\bR}{\bR} \tilde{\bSig}_k \Kab{\bR}{\bR}  $,
	respectively.
	This parametrization constitutes a computational shortcut, since the basis functions are very easy to interpret and do not include any matrix multiplication. 
	Note that also the log marginal likelihood discussed below is not affected by this transformation.

	\subsection{Prediction}
	Given a new $\bX_* \in \R^{A\times D}$,  
	the predictive distribution after seeing $\by_{1:k}$ of the sparse GP methods can be obtained by
	\begin{align}
	\label{eq:batch_pred}
	\begin{split}
	\pc{\bff_*}{\by_{1:k}}
	&=
	\int \pc{\bff_*}{\bu} \pc{\bu}{\by_{1:k}} \diff \bu
	\\
	&= \NN{\bff_*}{\bH_*\bmu_k, \bH_* \bSig_k \bH_{*}^{T} + \bV_*}
	\end{split}
	\end{align}
	using $\pc{\bff_*}{\bu}=\NN{\bff_*}{\bH_*\bu,\bV_*}$ with 
	$\bH_* = \Kab{*}{\bR}  \Kab{\bR}{\bR}^{-1}$
	and $\bV_*$ the model specific prediction covariance. 
	The predictions for $\by^*$ are obtained by adding $\sign\II$ to the covariance of $\bff_*\vert \by_{1:k} $. 
	At step $K$, by applying the Woodbury identity 
	to the batch covariance 
	$\bSig_K$ in \eqref{eq:batch}, we get for the predictive distribution in \eqref{eq:batch_pred} 
	\begin{align}
	\label{eq:batch_pred_form}
	\bmu_K^* &=  \bH_* \bSig_0 \bH^T
	\bSig \by,
	\\
	\bSig_K^* &= \bH_* \bSig_0 \bH_*^{T} - \bH_* \bSig_0 \bH^T
	\bSig
	\bH\bSig_0  \bH_*^{T}
	+\bV_*, \nonumber
	\end{align}
	where
	$\bSig = \left(
	\bH\bSig_0\bH^T +\bV
	\right)^{-1}$.
	Inserting the particular choices for 
	$\bSig_0$,
	$\bH $
	and
	$\bH_*$
	yields the usual formulation for the sparse predictive distribution
	\begin{align}
	\begin{split}
	\bmu_K^* &=  \Qab{*}{\bX}
	\left(
	\Qab{\bX}{\bX} + \bV
	\right)^{-1} \by;
	\\
	\bSig_K^* &=\Qab{*}{*} -  \Qab{*}{\bX}
	\left(
	\Qab{\bX}{\bX} + \bV
	\right)^{-1}
	\Qab{\bX}{*} + \bV_*.
	\end{split}
	\end{align}
	Depending on the choice of the covariances $\overline{\bV}$ and $\bV_*$, we obtain for instance \eqref{eq:PEP_pred} for PEP, or the analogous predictions for 
	VFE and FITC, respectively.
	\begin{figure*}[htb!]
		\centering
		\includegraphics[width=0.9\textwidth]{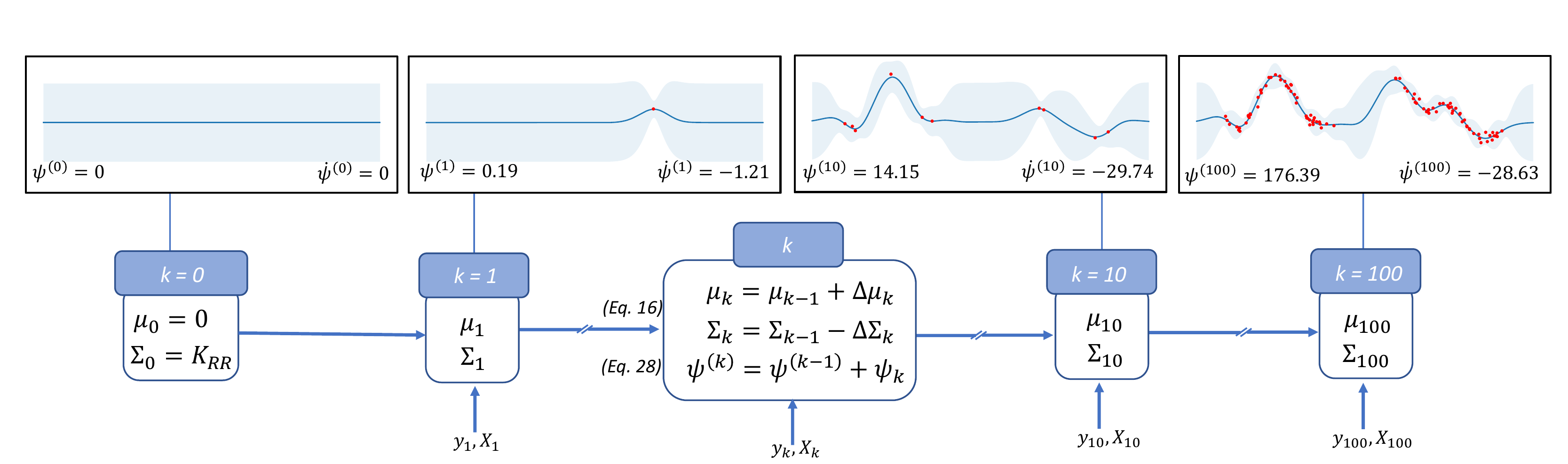}
		\caption{Online learning for the toy example in Sect. \ref{sse:online} for fixed $\bthe$ with batch size $B=1$.
			In each step $k$, the sample $\by_k, \bX_k$ is updated to the current posterior represented by $\bmu_k$ and $\bSig_k$ according to Eq. \eqref{eq:KF} together with the cumulative recursive bound $\psi^{(k)}$ (Eq. \eqref{eq:recursive_lik} and the numbers in the left corners of the plots). For recursive parameter estimation as discussed in Sect. \ref{sse:rec_grad_prop}, in addition to the posterior and the cumulative bound, the recursive derivatives are propagated for each $k$. The derivatives of this bound w.r.t.\ to the lengthscale are indicated in the right corners of the plots. 
			The cumulative bound and its derivative at step $k=100$ are equal to the corresponding batch version in Fig. \ref{fig_sparsevsfull}. 
		}
		\label{figKF}
	\end{figure*}

	\subsection{Online learning}
	\label{sse:online}
	This connection between sparse GP models and recursive estimation allows us to train the sparse GP models analytically online for streaming data for fixed $\bthe$.

	As an illustrative example consider $N=100$ data samples in $D=1$, we are interested in training a PEP model with $\alpha=0.5$ with $M=15$ inducing points with fixed $\bthe$. %
	Let's assume that the data samples $\{\bx_k, y_k\}$ arrive sequentially in  a stream. Thus we have $B=1$ and $K=100$. Here we use the  transformation and we apply, for each data sample $k$, the recursion in \eqref{eq:KF}. We note that here $\tilde{\br}_k$, $\tilde{\bV}_k$ and $\tilde{\bS}_k$ are numbers as $B=1$ and $\tilde{\bH}_k, \tilde{\bG}_k \in \R^{15}$.

	We obtain predictions for new data $\bX_*$, by applying \eqref{eq:batch_pred} with $\tilde{\bH}_{*} = \Kab{*}{\bR}$ and $\bV_* = \Kab{*}{\bR}  \Kab{\bR}{\bR}^{-1}  \Kab{\bR}{*}$. Note that there is no need to transform back the posterior over the inducing points, since it is already taken into account in the prediction step.
	After processing all  $N$ samples, the predictive distribution and the cumulative bound of log marginal likelihood correspond to the batch version, as shown in Fig. \ref{figKF}.

	\subsection{Marginal Likelihood}
	In the batch setting, the log marginal likelihood  $\log\p{\by}$ 
	can be computed by marginalizing out $\bu$, that is
	\begin{align}
	\label{eq:batch_marg_lik}
	\begin{split}
	\log\p{\by}
	&=
	\log \int \pc{\by}{\bu} \p{\bu} \diff \bu \\
	&= \log\NN{\by}{\bO, \bH\bSig_0\bH^T + \bV}.
	\end{split}
	\end{align}
	In the recursive setting, $ \p{\by}$ can be factorized into
	$
	\prod_{k=1}^K \pc{\by_k}{\by_{1:k-1}},
	$
	where 
	\begin{align}
	\label{eq:rec_marg}
	\pc{\by_k}{\by_{1:k-1}} 
	&= \NN{\by_k}{\bH_k \bmu_{k-1}, \bH_k\bSig_{k-1}\bH_k^T + \bV_k}
	\nonumber\\
	&= \NN{\br_k}{\bO, \bS_k}.
	\end{align}
	The log of the joint marginal likelihood involving all terms of \eqref{eq:rec_marg} can be explicitly written as
	\begin{align}
	\label{eq:rec_likelihood}
	\begin{split}
	&\log \prod_{k=1}^K \pc{\by_k}{\by_{1:k-1}}
	= \sum_{k=1}^K  \log \NN{\br_k}{\bO, \bS_k}
	\\
	&= -\frac{N}{2}\log 2\pi -\frac{1}{2}
	\sum_{k=1}^K 
	\log\deT{\bS_k} + \br_k^T \bS_k^{-1} \br_k.
	\end{split}
	\end{align}
	The iterative maximization of a lower bound  of the recursive factorized marginal likelihood in \eqref{eq:rec_marg} leads to the recursive KF updates 
	in \eqref{eq:KF} for the posterior and to the lower bound 
	\begin{align}
	\label{eq:recursive_lik0}
	\begin{split}
	\psi(\bthe)
	= \sum_{k=1}^K  \log \NN{\br_k^{\bthe}}{\bO, \bS_k^{\bthe}} - a_k(\bthe)
	\end{split}
	\end{align}
	which includes a model specific regularization term $a_k$ (see the right-most column in Fig. \ref{fig:parameters}). 
	We refer to this as the \textit{recursive collapsed bound} and
	a detailed derivation 
	for the VFE model is given in App.~\ref{se:alt_view}. Using the model specific quantities $\overline{\bs{V}}_k$ and $a_k$, this recursive computation of the lower bound of the marginal likelihood are equivalent to the batch counterparts for all sparse models, for instance
	\eqref{eq:collapsed_bound}  and \eqref{eq:PEP_log} for VFE and PEP, respectively.

	\section{Hyper-parameters Estimation}
	\label{se:vary_hypers}
	
	The previous section presented an online procedure for training sparse GP models at fixed hyper-parameters $\bthe$. Here we show that, by exploiting the connections highlighted before, we can optimize $\bthe$ sequentially. The \textit{recursive collapsed bound} in \eqref{eq:recursive_lik0} decomposes into a recursive sum over the mini-batches which allows to optimize the hyper-parameters $\bthe$ sequentially as opposed to the collapsed bound \eqref{eq:collapsed_bound}. Our bound \eqref{eq:recursive_lik0} enables the application of stochastic optimization without needing to estimate all entries in the posterior mean vector and covariance matrix as in SVGP. Compared to the uncollapsed bound in \eqref{eq:uncollapsed_bound}, the variational distribution is recursively and analytically eliminated, thus reducing the number of parameters to be numerically estimated drastically from $\mathcal{O}(MD + M^2)$ to
	$\mathcal{O}(MD)$.
	
	Finding a 
	maximizer
	$\bthe \in \bThe$ of an objective function $\Psi(\bthe) = \sum_{k=1}^K \psi_k(\bthe)$ can be achieved by applying \textit{Stochastic Gradient Descent}(SGD), with the update 
	\begin{align}
	\label{eq:SGD}
	\bthe^{(t)} = \bthe^{(t-1)} - \gamma^{(t-1)} \parD{\psi_k}{\bthe}_{\vert \bthe=\bthe^{(t-1)}},
	\end{align}
	where $ \gamma^{(t-1)}$ might be a sophisticated function of $\bthe^{(0)},\ldots,\bthe^{(t-1)}$ (for instance using ADAM \cite{kingma2014adam}, where also a bias correction term is included).
	We call one pass over the $K$ mini-batches an epoch. We denote $\bthe^{(e,k)} \in \bThe^{(e,k)}$ the estimate of $\bthe$ in epoch $e\in E$ for mini-batch $k$.

	\subsection{Recursive Gradient Propagation (RGP)}
	\label{sse:rec_grad_prop}
	We rewrite the \textit{recursive collapsed bound} in
	\eqref{eq:recursive_lik0} as
	\begin{align}
	\label{eq:recursive_lik}
	\psi^{(K)}(\bthe)
	=\sum_{k=1}^K d_k(\bthe) - a_k(\bthe)
	= \sum_{k=1}^K \psi_k(\bthe)
	\end{align}
	where 
	$d_k(\bthe)= \log \NN{\br_k^{\bthe}}{\bO, \bS_k^{\bthe}}$ and $a_k(\bthe)$ the model specific regularization term. 
	Since $\psi^{(K)}(\bthe)$ decomposes into a (recursive) sum over the mini-batches, we directly compute the derivative of $\psi_k(\bthe)$ w.r.t.\ $\bthe\in \bThe$.
	The derivative of $a_k$ is straightforward, for $d_k$ we have
	\begin{align}
	\label{eq:dk}
	\parD{d_k(\bthe)}{\bthe}
	=-\frac{1}{2}\parD{\log\deT{\bS_k^{\bthe}}}{\bthe}
	-\frac{1}{2}\parD{
		(\br_k^{\bthe})^T (\bS_k^{\bthe})^{-1}\br_k^{\bthe}
	}{\bthe}
	\end{align}
	with
	$\br_k^{\bthe} = \by_k - \bH_k^{\bthe} \bmu_{k-1}^{\bthe}$ 
	and $\bS_k^{\bthe} = \bH_k^{\bthe}  \bSig_{k-1}^{\bthe} (\bH_k^{\bthe})^T + \bV_k^{\bthe}$. 
	It is important to note that ignoring naively the dependency of $\bthe$ through $\bmu_{k-1}^{\bthe}$ and $\bSig_{k-1}^{\bthe}$ completely forgets the past and thus results in overfitting the current mini-batch. 
	In order to compute the derivatives of $\bmu_{k}^{\bthe}$ and $\bSig_{k}^{\bthe}$, we exploit the chain rule for derivatives and recursively propagate the gradients of the mean and the covariance over time, that is
	\begin{align}
	\label{eq:KF_grad}
	\parD{\bu_k}{\theta}
	&=
	\parD{\bu_{k-1}}{\theta}
	+
	\parD{\bG_{k}}{\theta} \br_k
	+
	\bG_{k} \parD{\br_{k}}{\theta}
	\\
	\parD{\bSig_k}{\theta}
	&=
	\parD{\bSig_{k-1}}{\theta}
	-
	\parD{\bG_{k}}{\theta} \bS_k \bG_{k}^T
	-
	\bG_{k}\parD{\bS_{k}}{\theta} \bG_{k}
	-
	\bG_{k} \bS_k \parD{\bG_{k}^T}{\theta} 
	,\nonumber
	\end{align}
	where $\parD{\bG_{k}}{\theta}$, $\parD{\br_{k}}{\theta}$ and
	$\parD{\bS_{k}}{\theta}$ are computed recursively according to \eqref{eq:KF}. 
	Computing the derivatives of $d_k$ as explained in \eqref{eq:dk} and \eqref{eq:KF_grad}, the stochastic gradient 
	\begin{align}
	\label{eq:dpsi}
	\parD{\psi_k(\bthe)}{\theta} = \parD{d_k(\bthe)}{\theta} - \parD{a_k(\bthe)}{\theta}
	\end{align}
	can be computed for each mini-batch $k$.
	
	\begin{proposition} 
		\label{prop:gradients}
		Consider a fixed parameter vector $\bar{\bthe} \in \bThe$, the gradients of the full batch lower bound $\LL_{PEP}(\bar{\bthe})$ in \eqref{eq:PEP_log} with respect to $\bthe$ are equal to the cumulative partial derivatives $\parD{\psi^{(K)}}{\bthe}$ of the recursive collapsed bound with the above learning procedure. That is, it holds
		$
		\parD{\LL_{PEP}(\bar{\bthe})}{\bthe}=\parD{\psi^{(K)}(\bar{\bthe})}{\bthe}.
		$
	\end{proposition}
	This shows that, when the gradients are cumulated over all data samples, each gradient step of our recursive procedure is equivalent to a gradient step for $\LL_{PEP}$ and, therefore, follows the gradients of an optimal collapsed lower bound. 
	
	The toy example in Fig. \ref{figKF} also shows this equivalence. The numbers in the bottom left and right corners show the cumulative recursive collapsed bound $\psi^{(k)}$ and its cumulative derivative $\parD{\psi^{(k)}}{\l}$ (abbreviated as $\dot{\psi}^{(k)}$) with respect to the lengthscale. The lower bound of the marginal likelihood as well as its derivatives are exactly the same value
	as the corresponding batch counterpart in Fig. \ref{fig_sparsevsfull}.
	
	For \textit{Stochastic Recursive Gradient Propagation} (SRGP),
	in each epoch $e$ and mini-batch $k$, we interleave the update step of the inducing points in Eq.  \eqref{eq:rec} with the SGD update \eqref{eq:SGD} of the parameters $\bthe^{(e,k)}$, i.e. 
	\[
	\pc{\bu}{\by_{1:k}, \bthe^{(e,k)}} 
	\approxprop
	\pc{\by_k}{\bu , \bthe^{(e,k)}}
	\pc{\bu}{\by_{1:k-1}, \bthe^{(e,k-1)}} .
	\]
	More concretely, we update after each mini-batch $k$ the parameters $\bthe^{(e,k)}$ with \eqref{eq:dpsi},\eqref{eq:SGD} and propagate recursively the posterior with \eqref{eq:KF} and its derivative \eqref{eq:KF_grad}.
	In order to compute all the derivatives with respect to $\bthe \in \bThe$, we exploit several matrix derivative rules which simplify the computation significantly, see App.~\ref{se:details_rec_prop}. Finally note that the form of the gradients in eq.~\eqref{eq:dk} and~\eqref{eq:dpsi} implies that the noise in the stochastic gradient at step $k$ depends on the noise at step $k$-1, thus excluding standard convergence proofs such as \cite{Bottou2018}. Recent results on non-convex optimization problems \cite{Chen2019,Zhou2018,Zou2019} show convergence proofs for function classes that include our objective, nonetheless the Markov noise of our problem still excludes it from this general theory. For this reason we leave a convergence analysis to future work.
	
	\subsection{Computational complexity}
	
\begin{figure*}[htp!]
		\centering
		\includegraphics[width=.95\textwidth]{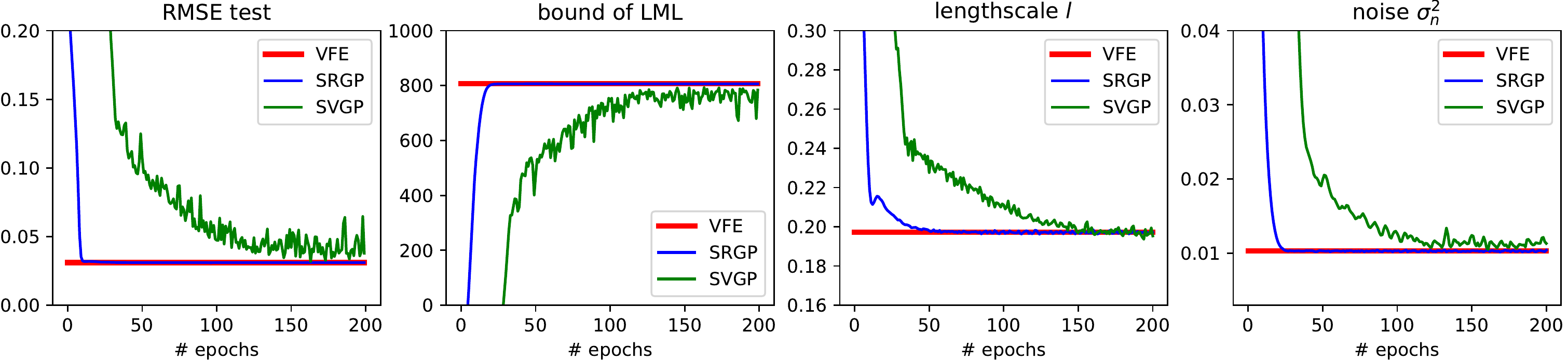}
		\caption{Convergence of SRGP (blue) on a 1-D toy to batch version VFE (red). Compared to SVGP (green), the convergence of the root-mean-squared-error (RMSE) of test points, the bound of the log marginal likelihood (LML) as well as the hyper-parameters is faster and more accurate.
			%
		}
		\label{fig:conv}
	\end{figure*}

	In the following, we assume that the batch size $B$ is larger than the number of inducing points $M$. For one mini-batch, the time complexity to update the posterior  is dominated by 
	matrix multiplications of size $B$ and $M$, thus 
	$\mathcal{O}(B^2M)$.
	In order to propagate the gradients of the posterior and to compute the derivative of the bound needs $\mathcal{O}(BM^2  )$ for a mini-batch and a parameter $\bthe \in \bThe$.   Thus, updating a mini-batch including all $\mathcal{O}(MD)$ parameters costs $\mathcal{O}(BM^3D +B^2M)$ for the SRGP method.
	Since SRGP stores the gradients of the posterior, it requires $\mathcal{O}(M^3D+BM)$ storage. 
	\\
	On the other hand, SVGP needs $\mathcal{O}(M^2+BM)$ storage and $\mathcal{O}(BM^3+B^2M)$ time per mini-batch, where the latter can be broken down into once $\mathcal{O}(B^2M)$ and $\mathcal{O}(BM)$ for each of the $\mathcal{O}(M^2)$ parameters.
	This means,  for moderate dimensions, our algorithm has the same time complexity as state of the art method SVGP. However, due to the analytic updates of the posterior we achieve an higher accuracy and less epochs are needed as shown in Fig. \ref{fig:conv} and in Sect. \ref{se:experiments} empirically.

	Fig. \ref{fig:conv} shows the convergence of SRGP on a 1-D toy example with $N=1000$ data samples and $M=15$ inducing points. The parameters are sequentially optimized with our recursive approach (blue) and as comparison with SVGP (green) with a mini-batch size of $B=100$ over several epochs. The root-mean-squared-error (RMSE) computed on test points, the bound of the log marginal likelihood (LML) as well as the hyper-parameters converge in a few iterations to the corresponding batch values of VFE (red). Due to the analytic updates of the posterior, the accuracy is higher and SRGP needs much less epochs until convergence.
	\subsection{Mini-batch size}
	
\begin{figure}[htp!]
		\centering
		\includegraphics[width=.35\textwidth]{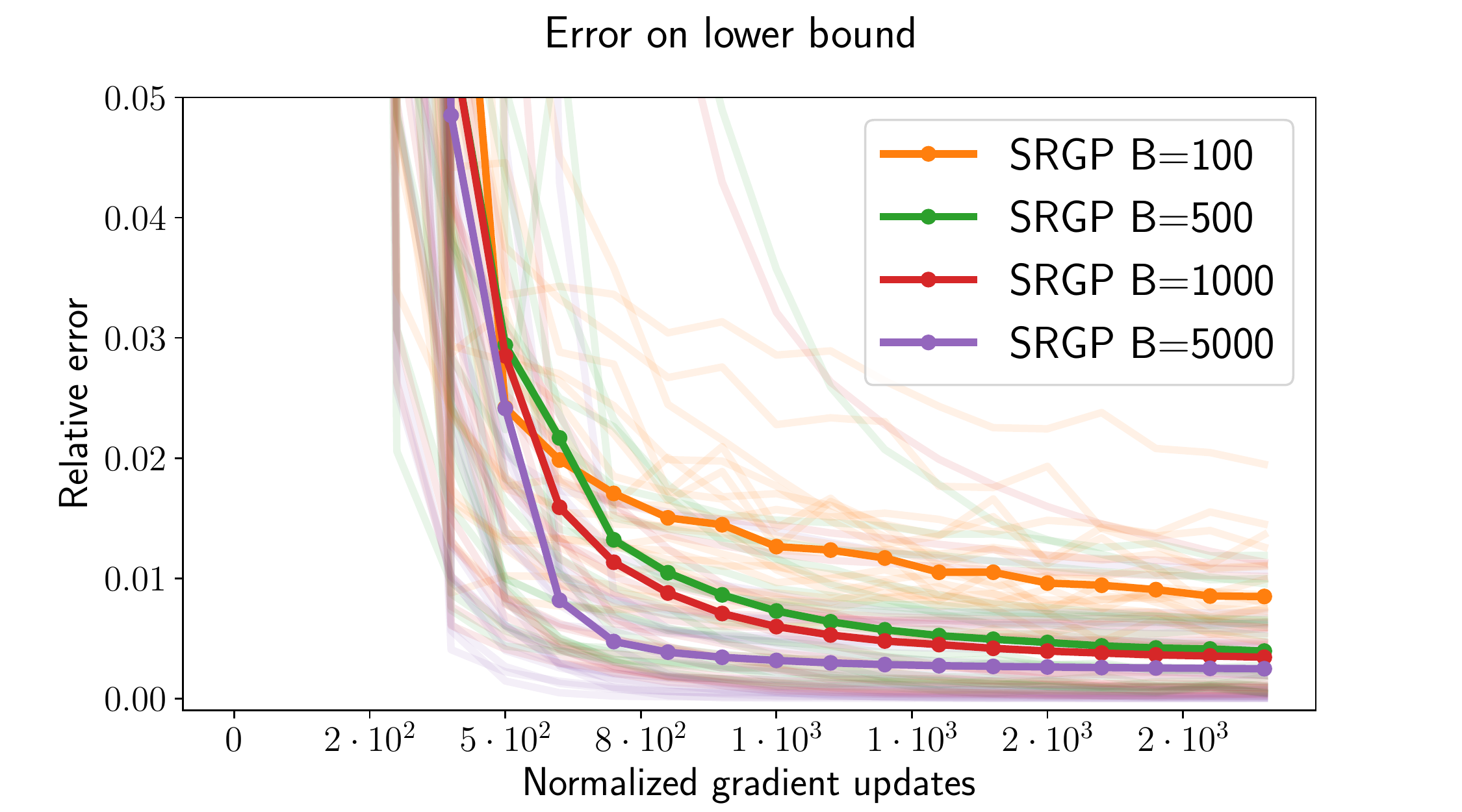}
		\caption{Relative error in the bound of the log marginal likelihood between full batch and SRGP. Average over $20$ repetitions.
			%
		}
		\label{fig:miniBatch}
	\end{figure}

	The size of the mini-batches has an impact on the speed of convergence of the algorithm. Proposition~\ref{prop:gradients} tells us that if we use a full batch, our algorithm requires the same number of gradient updates as a full batch method to converge. On the other hand smaller batches should require more updates and should lead to a higher variance in the results. Fig.~\ref{fig:miniBatch} shows a comparison of different mini-batch sizes on a 1-D toy example with $N=10'000$ data samples generated with the same parameters as in Sect.~\ref{sse:GPsims}. The convergence to the full batch value is slower as the batch size decreases. Moreover the variance of the error, over the repetitions, is much larger for smaller batch sizes: in the last 10 normalized gradient updates, the standard deviation of the error is on average $3.8 \times 10^{-3}$ for $B=100$ and $8.1 \times 10^{-4}$ for $B=5'000$, denoting a more stable procedure for higher batch sizes. As the mini-batch size increases the computational cost for each gradient update also increases. In this example one gradient update requires on average $4.9 \times 10^{-3}$ sec and $5.8 \times 10^{-2}$ sec with $B=100$ and $B=5'000$ respectively.\footnote{The times are measured on a laptop with a Intel i5-7300U CPU $@$ 2.6 GHz.} These considerations suggest that a reasonable choice is a large mini-batch size within the computational and time budgets.
	
	\section{Experiments}
	\label{se:experiments}
	%
	We first benchmark our method with $N=100'000$ synthetic data samples generated by a GP in several dimensions. Next, we apply our approach to the Airline data used in \cite{hensman2013gaussian} with a million of data samples. 
	Finally a more realistic setup is presented where we use up to a million data samples to train a nonlinear plant. We compare our SRGP method to full GP and sparse batch method VFE for a subset of data (using the implementation in GPy \cite{gpy2014}) and to the state of the art stochastic parameter estimation method SVGP implemented in GPflow \cite{GPflow2017}.
	Our algorithm works also for many other sparse models, however, only large-scale implementations of standard SVGP are  available (corresponding to the VFE model), thus we restrict the investigation to this model.
	
	\subsection{GP Simulation}
	\label{sse:GPsims}
	In this section we test our proposed learning procedure  
	on simulated GP data. We generate $N=100'000$ data samples from a zero-mean (sparse) GP with SE covariance kernel with hyper-parameters $\sigma_0 = 1, \sigma_n = 0.1$ and $l=\{0.1,0.2,0.5\}$ in $D=\{1, 2, 5\}$ dimensions.
	The initial $M=\{20,50,100\}$ inducing points are randomly selected points from the data and the hyper-parameters of a SE kernel with individual lengthscales for each dimension are initialized to the same values for both algorithms ($\sigma_0 = 1, \sigma_n = 1, l_1,\ldots,l_D=1$). All parameters are sequentially optimized with our recursive approach and with SVGP with a mini-batch size of $B=5000$. The stochastic gradient descent method ADAM \cite{kingma2014adam} is employed for both methods with learning rates $\{0.001, 0.005, 0.005\}$ for SVGP and  $\{0.0001, 0.001, 0.005\}$ for SRGP (based on some preliminary experiments). Each experiment is replicated $10$ times. 
	
\begin{figure}[htp!]
		\centering
		\includegraphics[width=.47\textwidth]{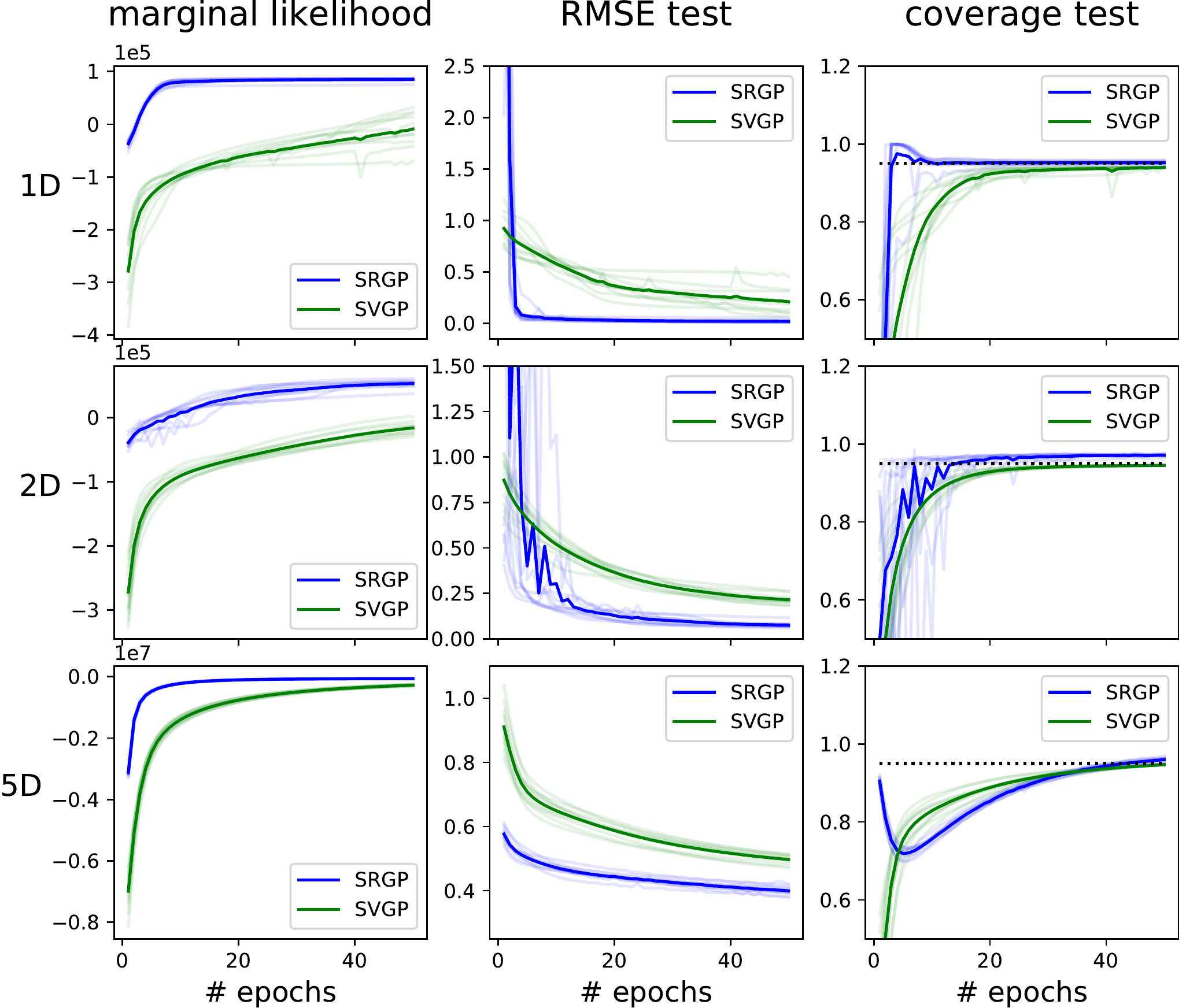}
		\caption{Convergence over $50$ epochs for $N=100'000$ synthetic GP data samples in several dimensions obtained by SVGP and our proposed method SRGP.
			%
			%
			%
			%
		}
		\label{fig:simGP}
	\end{figure}

	Fig. \ref{fig:simGP} shows the bound to the log marginal likelihood, the RMSE and the coverage of $10'000$ test points for the data dimensions $D=\{1, 2, 5\}$ of both methods over $50$ epochs. The shaded lines indicates the $10$ repetitions and the thick line correspond to the mean.   
	The recursive propagation of the gradients achieves faster convergence and more accurate performance regarding mean RMSE and smaller values for the log marginal likelihood. 
	The higher accuracy and faster convergence can possibly be explained by the analytic updates of the posterior mean and covariance which leads to less parameters to be optimized numerically.

	\subsection{Airline Data}
\begin{figure}[htp!]
		\centering
		\includegraphics[width=0.48\textwidth]{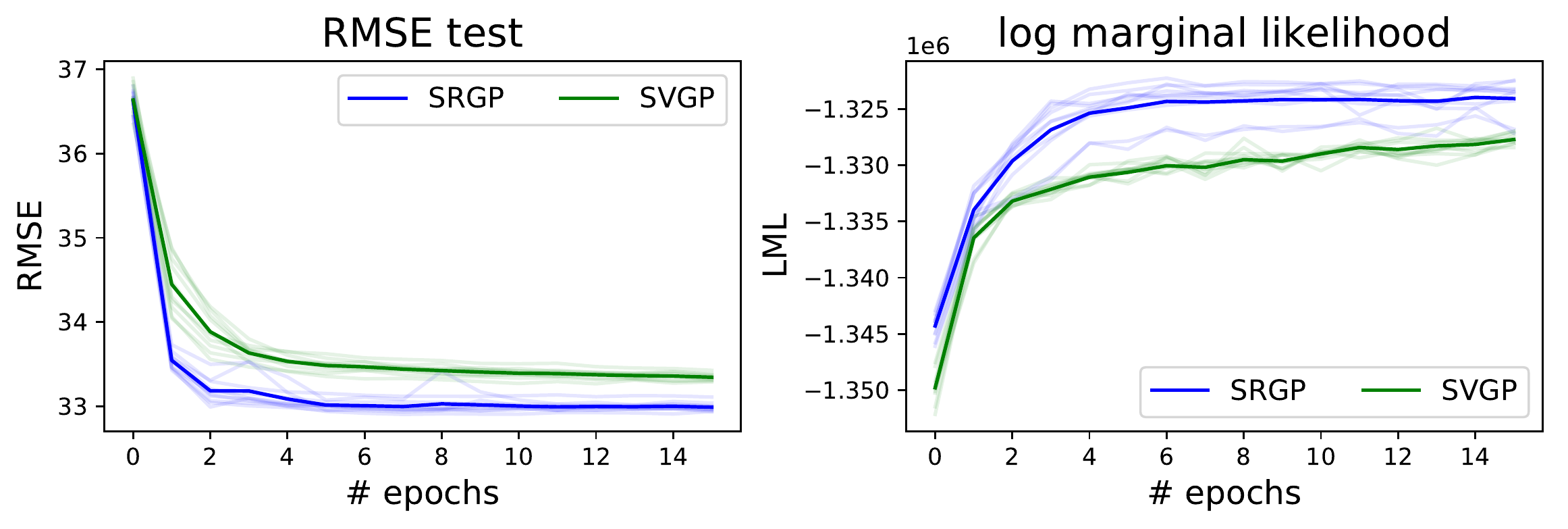}
		\caption{Convergence over several epochs of RMSE and bound to log marginal likelihood for $N=1'000'000$ samples from the Airline data for SRGP and SVGP.}
		\label{fig:airlineBig}
	\end{figure}

	For the second example we apply our recursive method to the Airline Data used in \cite{hensman2013gaussian}. It consists of flight arrival and departure times for more than 2 millions flights in the USA from January 2008 and April 2008. We preprocessed the data as similar as possible as described in \cite{hensman2013gaussian} resulting in 8 variables: age of the aircraft, distance that needs to be covered, airtime, departure time, arrival time, day of the week, day of the month and month. We trained our recursive method as well as SVGP with an SE kernel on $N=1'000'000$ data samples with $M=500$ inducing points randomly selected from the data and a mini-batch size of $B=10'000$. The ADAM learning rates are set to $0.005$ for both methods and the size of the test set is $50'000$. For $5$ different repetitions, the RMSE as a function of epochs is depicted in Fig. \ref{fig:airlineBig}. The mean coverage on test data (at $95\%$) is comparable for both methods with values of $0.92$ and $0.97$ for SVGP and SRGP respectively. The overall performance  of SRGP is superior to SVGP. 

	\subsection{Non-Linear Plant}
	GPs are a powerful way to model complex functions in a non-parametric way, thus they are suitable to learn the complex input output behavior of a non-linear plant. However, with full or even sparse batch GP methods the use is restricted to a few thousands of samples. With our sequential learning method, we are able to exploit the huge amount of available data by training with up to a million of samples.
	
\begin{figure}[htp!]
		\centering
		\includegraphics[width=0.48\textwidth]{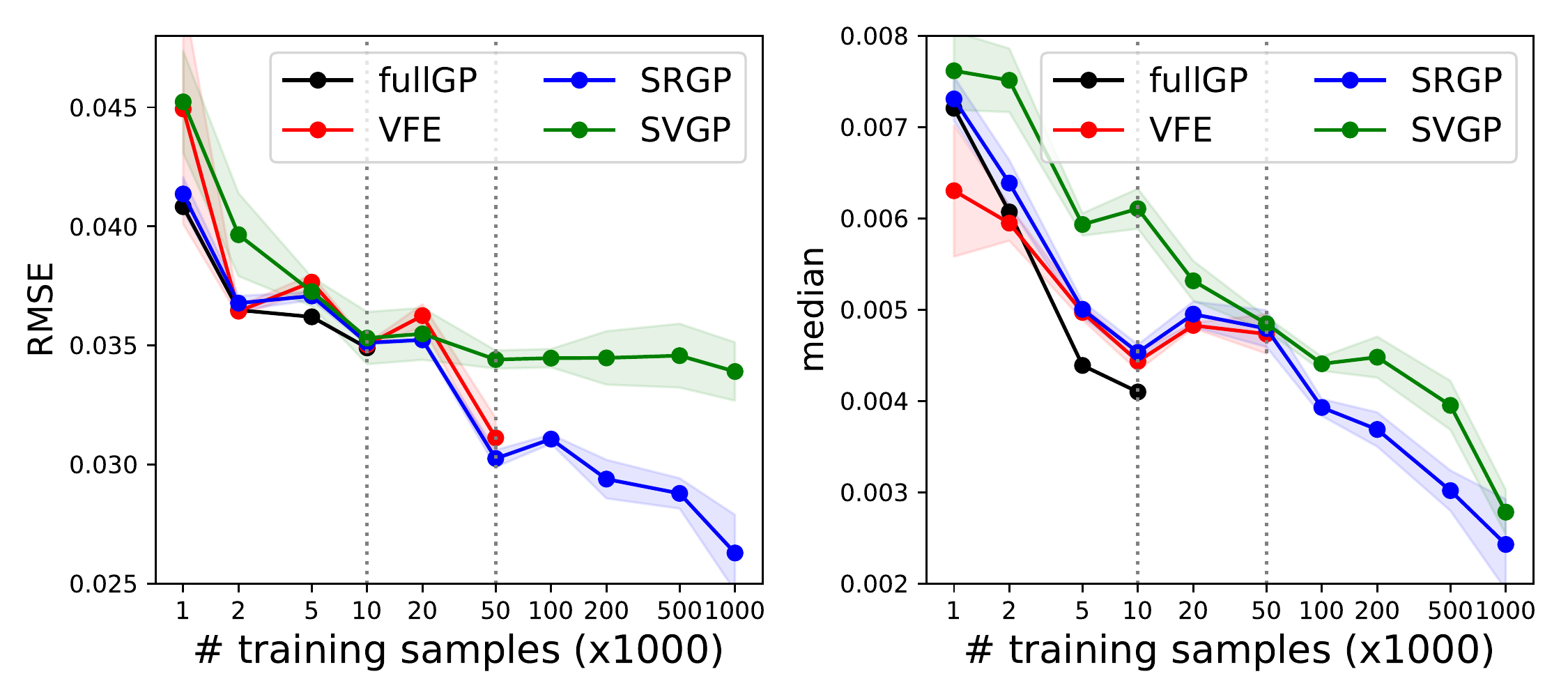}
		\caption{RMSE and median for full GP, batch sparse GP (VFE), sequential SVGP, and our recursive method (SRGP) trained on varying time horizons (logarithmic scale).
			The grey dotted vertical lines at $10'000$ and $50'000$ indicate the maximal samples used for training full GP and sparse batch GP (VFE), respectively.} 
		\label{fig:application_results}
	\end{figure}

	We consider a Continuous Stirred Tank Reactor (CSTR). The dynamic model of the plant is
	{\small
	\begin{align}
	\nonumber
	&\tfrac{d}{dt}h(t)=w_1(t)+w_2(t)-0.2\sqrt{h(t)}\\ 
	\nonumber
	&\tfrac{d}{dt} C_b(t)=(C_{b1}-C_{b}(t))\tfrac{w_1(t)}{h(t)}+(C_{b2}-C_b(t))\tfrac{w_2(t)}{h(t)}-\tfrac{k_1C_b(t)}{(1+k_2C_b(t))^2},
	\end{align}}
	where $C_b(t)$ is the product concentration at the output of the process, $h(t)$ is the liquid level, $w_1(t)$ is the flow rate of concentrated feed $C_{b1}$, and $w_2(t)$ is the flow rate of the diluted feed $C_{b2}$. The input concentrations are $C_{b1} = 24.9$ and $C_{b2} = 0.1$. The constants associated with the rate of consumption are $k_1 = k_2 = 1$.
	The objective of the controller is to maintain the product concentration by changing the flow $w_1(t)$. To simplify the example, we assume that $w_2(t) = 0.1$ and that the level of the tank $h(t)$ is not controlled.
	We denote the controlled outputs $C_b(t),C_b(t-1),\dots,C_b(t-p)$ as $f_{t},f_{t-1},\ldots,f_{t-p}$ and the control variables
	as $w_t,w_{t-1},\ldots,w_{t-p}$. 
	Therefore, the plant identification problem can be shaped into the problem of estimating the non-linear function $f_t = g(f_{t-1},\ldots,f_{t-p},w_t,w_{t-1},\ldots,w_{t-p})$ which depends on the $p$ previous values as well as on the current and the $p$ past control values $w$. 
	\begin{figure}[htp!]
		\centering
		\includegraphics[width=0.46\textwidth]{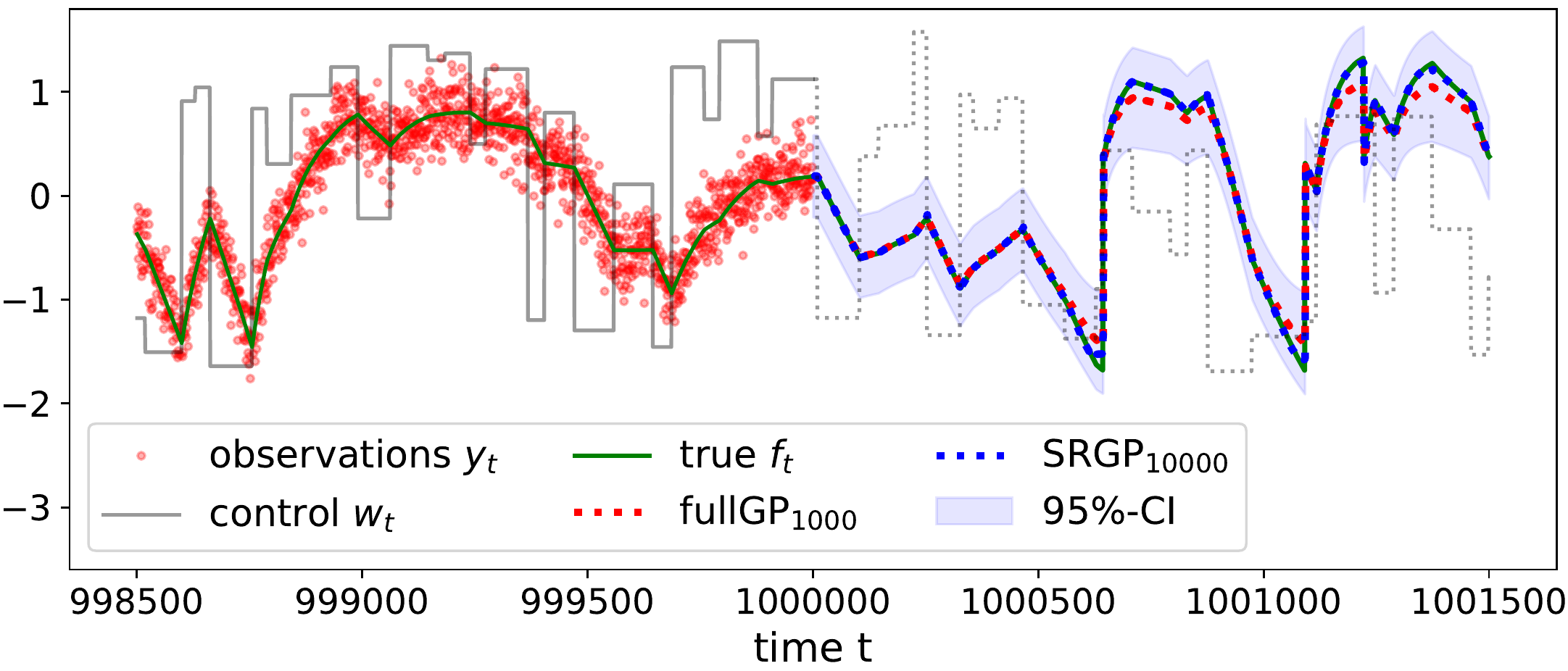}
		\caption{Training and prediction phases for non-linear plant. }
		\label{fig:application_setup}
	\end{figure}
	However, we can only observe a noisy version of the controlled response, that is $y_t = f_t + \varepsilon_t$ with $\varepsilon\sim \NNo{0,\sigma_n^2}$.  
	Using a sampling rate of $0.2s$, we have generated $1'200'000$ observations (about $3$ days of observations).
	The plant input is a series of steps, with random height (in the interval  $[0,4]$), occurring at random intervals (in the interval  $[5,20]s$).
	For different numbers $N_{train}$, we use the samples $y_{10^6-N_{train}},\ldots,y_{10^6}$ for training and the last $200'000$ are used as a test set.
	The goal is to learn a model for the controlled response $y_t$ given $\bx_t = [y_{t-1},y_{t-2},w_t,w_{t-1},w_{t-2}]^T \in \R^5$ for the particular choice of $p=2$.  We model the non-linear function $g$ with a GP with a SE kernel. For comparison, we train full GP and sparse batch GP (with $100$ inducing points) on a time horizon $N_{train}$ of up to $10'000$ and $50'000$ past values, respectively. 
	With the sequential version SVGP and our recursive gradient propagation method SRGP (both with $100$ inducing points and mini-batch size of $1'000$), we use a time horizon of up to a million.
	This situation is depicted in Fig. \ref{fig:application_setup}, where for $1500$ training samples $y_t$ (red dots), the true (unknown) function $f_t$ (green) and the control input $w_t$ (grey) is shown together with the predicted values with full GP (red dotted) and recursive GP (blue dotted) trained on a time horizon of $1'000$ and $10'000$, respectively.
	In Fig. \ref{fig:application_results}, the RMSE and the median computed on the test set (with $10$ repetitions) is depicted for full GP, sparse GP (VFE), SVGP and SRGP trained with varying time horizons. For small and medium training sizes, when the batch methods are applicable, our recursive method achieves the same performance as the batch counterpart (VFE) and is comparable to full GP. Due to the analytic updates of the posterior, SRGP outperforms SVGP regarding both RMSE and median for all training sizes. By exploiting more than several thousand past values, a significant increase in performance of SRGP can be still observed, thus it constitutes an approach to accurately scale GPs up to a million of past values.

	\section{Conclusion}
	\label{se:conclusion}
	
	In this paper we introduced a recursive inference and parameter estimation method SRGP for a general class of sparse GP approximations. Since the posterior updates are given analytically, one pass through the data is sufficient to compute the posterior for given parameters.
	For parameter estimation, we proposed a recursive collapsed bound to the log marginal likelihood that matches exactly the batch version but can be used for stochastic estimation.
	Due to the analytic updates of the posterior our method has much less parameters to be estimated numerically. As a consequence, the experimental section  showed that our recursive method needs less epochs and has superior accuracy compared to state of the art, thus constitutes an efficient methodology for scaling GPs to big data problems.
	\\
	Our approach could be enhanced in several directions.
	While the proposed method only exploits the update equations of the KF, an interesting direction would be to include a dynamic in a state space model that takes into account the varying hyper-parameters which makes it also applicable for the streaming setting as \cite{bui2017streaming}.
	Moreover, we further plan to investigate distributed parameter estimation based on an information filter formulation of the problem. 
	Finally, we aim to provide a convergence analysis of the proposed method  to recursively learn the hyper-parameters using a similar approach recently employed for 
	ADAM or AMSGrad \cite{Sashank_adam2018,TranPhong_AMSGrad2019}.
	
	\begin{ack}                               
		\noindent
		This work is supported by the Swiss National Research Programme 75 "Big Data" grant n.~407540\_167199~/~1.
	\end{ack}
	
	\bibliographystyle{plain}        
	\bibliography{ref}           

\begin{thebibliography}{10}

\bibitem{Barber2014GaussianPF}
David Barber and Yali Wang.
\newblock Gaussian processes for bayesian estimation in ordinary differential
  equations.
\newblock In {\em ICML}, 2014.

\bibitem{benavoli2016b}
Alessio Benavoli and Marco Zaffalon.
\newblock State space representation of non-stationary gaussian processes.
\newblock 2016.

\bibitem{Bijl2016OnlineSG}
Hildo Bijl, Thomas~B. Sch{\"o}n, Jan-Willem van Wingerden, and Michel
  Verhaegen.
\newblock Online sparse gaussian process training with input noise.
\newblock {\em CoRR}, abs/1601.08068, 2016.

\bibitem{Bishop2006}
Christopher~M. Bishop.
\newblock {\em Pattern Recognition and Machine Learning (Information Science
  and Statistics)}.
\newblock Springer-Verlag, Berlin, Heidelberg, 2006.

\bibitem{Bottou2018}
L{\'{e}}on Bottou, Frank~E. Curtis, and Jorge Nocedal.
\newblock {Optimization methods for large-scale machine learning}.
\newblock {\em SIAM Review}, 60(2):223--311, 2018.

\bibitem{bui2017streaming}
Thang~D Bui, Cuong Nguyen, and Richard~E Turner.
\newblock Streaming sparse gaussian process approximations.
\newblock In {\em Advances in Neural Information Processing Systems}, pages
  3301--3309, 2017.

\bibitem{bui2016unifying}
Thang~D Bui, Josiah Yan, and Richard~E Turner.
\newblock A unifying framework for sparse gaussian process approximation using
  power expectation propagation.
\newblock {\em Journal of Machine Learning Research}, 18:1--72, 2017.

\bibitem{carron2016machine}
Andrea Carron, Marco Todescato, Ruggero Carli, Luca Schenato, and Gianluigi
  Pillonetto.
\newblock Machine learning meets kalman filtering.
\newblock In {\em 2016 IEEE 55th Conference on Decision and Control (CDC)},
  pages 4594--4599. IEEE, 2016.

\bibitem{chen2012estimation}
Tianshi Chen, Henrik Ohlsson, and Lennart Ljung.
\newblock On the estimation of transfer functions, regularizations and gaussian
  processes—revisited.
\newblock {\em Automatica}, 48(8):1525--1535, 2012.

\bibitem{Chen2019}
Xiangyi Chen, Sijia Liu, Ruoyu Sun, and Mingyi Hong.
\newblock {On the convergence of a class of Adam-type algorithms for non-convex
  optimization}.
\newblock {\em 7th International Conference on Learning Representations, ICLR
  2019}, 2019.

\bibitem{csato2002sparse}
Lehel Csat{\'o} and Manfred Opper.
\newblock Sparse online gaussian processes.
\newblock {\em Neural computation}, 14(3):641--668, 2002.

\bibitem{frigola2013bayesian}
Roger Frigola, Fredrik Lindsten, Thomas~B Sch{\"o}n, and Carl~Edward Rasmussen.
\newblock Bayesian inference and learning in gaussian process state-space
  models with particle mcmc.
\newblock In {\em Advances in Neural Information Processing Systems}, pages
  3156--3164, 2013.

\bibitem{gpy2014}
{GPy}.
\newblock {GPy}: A gaussian process framework in python.
\newblock http://github.com/SheffieldML/GPy, since 2012.

\bibitem{hartikainen2010kalman}
Jouni Hartikainen and Simo S{\"a}rkk{\"a}.
\newblock Kalman filtering and smoothing solutions to temporal gaussian process
  regression models.
\newblock In {\em 2010 IEEE International Workshop on Machine Learning for
  Signal Processing}, pages 379--384. IEEE, 2010.

\bibitem{hensman2013gaussian}
James Hensman, Nicolo Fusi, and Neil~D Lawrence.
\newblock Gaussian processes for big data.
\newblock In {\em Conference for Uncertainty in Artificial Intelligence}, 2013.

\bibitem{hoang2015unifying}
Trong~Nghia Hoang, Quang~Minh Hoang, and Bryan Kian~Hsiang Low.
\newblock A unifying framework of anytime sparse gaussian process regression
  models with stochastic variational inference for big data.
\newblock In {\em ICML}, pages 569--578, 2015.

\bibitem{hoffman2013stochastic}
Matthew~D Hoffman, David~M Blei, Chong Wang, and John Paisley.
\newblock Stochastic variational inference.
\newblock {\em The Journal of Machine Learning Research}, 14(1):1303--1347,
  2013.

\bibitem{kingma2014adam}
Diederik~P Kingma and Jimmy Ba.
\newblock Adam: A method for stochastic optimization.
\newblock {\em arXiv preprint arXiv:1412.6980}, 2014.

\bibitem{kocijan2005dynamic}
Ju{\v{s}} Kocijan, Agathe Girard, Bla{\v{z}} Banko, and Roderick Murray-Smith.
\newblock Dynamic systems identification with gaussian processes.
\newblock {\em Mathematical and Computer Modelling of Dynamical Systems},
  11(4):411--424, 2005.

\bibitem{liu2018gaussian}
Haitao Liu, Yew-Soon Ong, Xiaobo Shen, and Jianfei Cai.
\newblock When gaussian process meets big data: A review of scalable gps.
\newblock {\em arXiv preprint arXiv:1807.01065}, 2018.

\bibitem{Macdonald2015}
Benn Macdonald, Catherine Higham, and Dirk Husmeier.
\newblock Controversy in mechanistic modelling with gaussian processes.
\newblock In {\em Proceedings of the 32Nd International Conference on
  International Conference on Machine Learning - Volume 37}, ICML'15, pages
  1539--1547. JMLR.org, 2015.

\bibitem{GPflow2017}
Alexander G. de~G. Matthews, Mark {van der Wilk}, Tom Nickson, Keisuke. Fujii,
  Alexis {Boukouvalas}, Pablo {Le{\'o}n-Villagr{\'a}}, Zoubin Ghahramani, and
  James Hensman.
\newblock {{GP}flow: A {G}aussian process library using {T}ensor{F}low}.
\newblock {\em Journal of Machine Learning Research}, 18(40):1--6, apr 2017.

\bibitem{mattosrecurrent16}
César Lincoln~C. Mattos, Zhenwen Dai, Andreas Damianou, Jeremy Forth,
  Guilherme~A. Barreto, and Neil~D. Lawrence.
\newblock Recurrent {G}aussian processes.
\newblock In Hugo Larochelle, Brian Kingsbury, and Samy Bengio, editors, {\em
  Proceedings of the International Conference on Learning Representations},
  volume~3, Caribe Hotel, San Juan, PR, 00 2016.

\bibitem{pillonetto2016new}
Gianluigi Pillonetto.
\newblock A new kernel-based approach to hybrid system identification.
\newblock {\em Automatica}, 70:21--31, 2016.

\bibitem{pillonetto2010new}
Gianluigi Pillonetto and Giuseppe De~Nicolao.
\newblock A new kernel-based approach for linear system identification.
\newblock {\em Automatica}, 46(1):81--93, 2010.

\bibitem{pillonetto2014kernel}
Gianluigi Pillonetto, Francesco Dinuzzo, Tianshi Chen, Giuseppe De~Nicolao, and
  Lennart Ljung.
\newblock Kernel methods in system identification, machine learning and
  function estimation: A survey.
\newblock {\em Automatica}, 50(3):657--682, 2014.

\bibitem{quinonero2005unifying}
Joaquin Qui{\~n}onero-Candela and Carl~Edward Rasmussen.
\newblock A unifying view of sparse approximate gaussian process regression.
\newblock {\em Journal of Machine Learning Research}, 6(Dec):1939--1959, 2005.

\bibitem{rasmussen2006gaussian}
Carl~Edward Rasmussen and Christopher~KI Williams.
\newblock {\em Gaussian processes for machine learning}, volume~1.
\newblock MIT press Cambridge, 2006.

\bibitem{Sashank_adam2018}
Sashank~J. Reddi, Satyen Kale, and Sanjiv Kumar.
\newblock On the convergence of adam and beyond.
\newblock In {\em International Conference on Learning Representations}, 2018.

\bibitem{sarkka2013spatiotemporal}
Simo Sarkka, Arno Solin, and Jouni Hartikainen.
\newblock Spatiotemporal learning via infinite-dimensional bayesian filtering
  and smoothing: A look at gaussian process regression through kalman
  filtering.
\newblock {\em IEEE Signal Processing Magazine}, 30(4):51--61, 2013.

\bibitem{seeger2003fast}
Matthias Seeger, Christopher Williams, and Neil Lawrence.
\newblock Fast forward selection to speed up sparse gaussian process
  regression.
\newblock In {\em Artificial Intelligence and Statistics 9}, number
  EPFL-CONF-161318, 2003.

\bibitem{silverman1985some}
Bernhard~W Silverman.
\newblock Some aspects of the spline smoothing approach to non-parametric
  regression curve fitting.
\newblock {\em Journal of the Royal Statistical Society. Series B
  (Methodological)}, pages 1--52, 1985.

\bibitem{smola2001sparse}
Alex~J Smola and Peter~L Bartlett.
\newblock Sparse greedy gaussian process regression.
\newblock In {\em Advances in neural information processing systems}, pages
  619--625, 2001.

\bibitem{snelson2006sparse}
Edward Snelson and Zoubin Ghahramani.
\newblock Sparse gaussian processes using pseudo-inputs.
\newblock In {\em Advances in Neural Information Processing Systems}, pages
  1257--1264, 2006.

\bibitem{svensson2017flexible}
Andreas Svensson and Thomas~B Sch{\"o}n.
\newblock A flexible state--space model for learning nonlinear dynamical
  systems.
\newblock {\em Automatica}, 80:189--199, 2017.

\bibitem{titsias2009variational}
Michalis Titsias.
\newblock Variational learning of inducing variables in sparse gaussian
  processes.
\newblock In {\em Artificial Intelligence and Statistics}, pages 567--574,
  2009.

\bibitem{todescato2017efficient}
Marco Todescato, Andrea Carron, Ruggero Carli, Gianluigi Pillonetto, and Luca
  Schenato.
\newblock Efficient spatio-temporal gaussian regression via kalman filtering.
\newblock {\em arXiv preprint arXiv:1705.01485}, 2017.

\bibitem{TranPhong_AMSGrad2019}
P.~T. {Tran} and L.~T. {Phong}.
\newblock On the convergence proof of amsgrad and a new version.
\newblock {\em IEEE Access}, 7:61706--61716, 2019.

\bibitem{wahba1999bias}
Grace Wahba, Xiwu Lin, Fangyu Gao, Dong Xiang, Ronald Klein, and Barbara Klein.
\newblock The bias-variance tradeoff and the randomized gacv.
\newblock In {\em Advances in Neural Information Processing Systems}, pages
  620--626, 1999.

\bibitem{Zhou2018}
Dongruo Zhou, Yiqi Tang, Ziyan Yang, Yuan Cao, and Quanquan Gu.
\newblock {On the Convergence of Adaptive Gradient Methods for Nonconvex
  Optimization}.
\newblock 2018.

\bibitem{Zou2019}
Fangyu Zou, Li~Shen, Zequn Jie, Weizhong Zhang, and Wei Liu.
\newblock {A sufficient condition for convergences of Adam and RMSProp}.
\newblock {\em Proceedings of the IEEE Computer Society Conference on Computer
  Vision and Pattern Recognition}, 2019-June(1):11119--11127, 2019.

\bibitem{Alvarez13}
M.~A. {Álvarez}, D.~{Luengo}, and N.~D. {Lawrence}.
\newblock Linear latent force models using gaussian processes.
\newblock {\em IEEE Transactions on Pattern Analysis and Machine Intelligence},
  35(11):2693--2705, Nov 2013.

\end{thebibliography}
	
	
	
	\appendix

	\section{Details for Recursive Gradient Propagation}
	\label{se:details_rec_prop}
	\allowdisplaybreaks
	We show here the computation for the recursive gradient propagation from Sect. \ref{sse:rec_grad_prop} for the PEP model. For other models, $a_k$ and $\overline{\bV}$ from the Table \ref{fig:parameters} could be used correspondingly.
	We will use the following notation:
	$\dIAG{\bA} = \bs{d}, d_i = a_{ii}$, $\DIAG{\bs{d}} = \bA, a_{ii} = d_i, a_{ij} = 0$, $\bA \odot \bB = \bC, c_{ij} = a_{ij} b_{ij}$, $\bA \div \bB = \bC, c_{ij} = \frac{a_{ij}}{ b_{ij}}$, $\bA^{\odot 2} = \bC, c_{ij} = a_{ij}^2$,
	$\dot{\bA} = \parD{\bA(\theta)}{\theta}, \forall \theta \in \bThe$, $\sumall{\bA} =  \sum_{i,j} a_{ij}$, $\mathrm{1}_{[z]} = 1 \text{ if } z=\text{true}, 0 \text{ other.}$

	\textbf{\textit{Initialization}}
	\vspace{-0.2cm}
	\begin{align*}
	\bet_0 &= \bs{0}; 
	&\dot{\bet}_0 = \bs{0}; \\
	\bLam_0&=\Kab{\bR}{\bR}^{-1}; 
	&\dot{\bLam}_0=-\Kab{\bR}{\bR}^{-1}\dot{\bs{K}}_{\bR\bR}\Kab{\bR}{\bR}^{-1}; \\ 
	\psi_0 &= -\frac{N}{2}\log2\pi;  
	&\dot{\psi}_0 = 0; \\
	\bSig_0&=\Kab{\bR}{\bR}; 
	&\text{logDet}_0 = \log\deT{\bLam_0}; 
	\end{align*}
	\vspace{-0.2cm}
	\textbf{\textit{Natural Mean and Precision Updates}}
	\vspace{-0.1cm}
	\begin{align*}
	\bH_k &= \Kab{\bX_k}{\bR}\Kab{\bR}{\bR}^{-1}; \\
	\bs{d}_k&=
	\dIAG{\Kab{\bX_k}{\bX_k} - \Kab{\bX_k}{\bR}\Kab{\bR}{\bR}^{-1}\Kab{\bR}{\bX_k}}
	;\\
	\bv_k&=\alpha \bs{d}_k +  \sigma_n^2 \mathbbm{1};
	\quad
	\bV_k^{-1}=\DIAG{\mathbbm{1}\div \bv_k};\\
	a_k&=
	\frac{1-\alpha}{\alpha} \left(
	\sum_{i=1}^{B}{ \log([\bv_k]_i)} - B\log\sigma_n^2
	\right);\\
	\\
	\br_k &= \by_k - \bH_k \bSig_{k-1} \bet_{k-1}; \\
	\bet_k &= \bet_{k-1} + \bH_k^T \bV_k^{-1}\by_k ; \\
	\bLam_k &= \bLam_{k-1} + \bH_k^T \bV_k^{-1} \bH_k ; \\
	\bSig_k, \log\deT{\bLam_k} &= \bLam_k^{-1}, \log\deT{\bLam_k};\\
	\bS_k^{-1} &= \bV_k^{-1}
	- \bV_k^{-1} \bH_k \bSig_k \bH_k^T \bV_k^{-1};\\
	\psi_k &= \psi_{k-1} -\frac{1}{2}\left(
	\log\deT{\bLam_k} - \log\deT{\bLam_{k-1}} 
	\right. \\ &\left.
	- \log\deT{\bV_k^{-1}}
	+ \br_k^T\bS_k^{-1}  \br_k  + a_k \right)
	\end{align*}

	\textbf{\textit{Intermediate Derivatives}}
	\begin{align*}
	\dot{L}_{d\bH_k}
	&=2  \left( \bV_k^{-1} \bH_k \bSig_k
	- \bS_k^{-1} \br_k ( \bSig_{k-1} \bet_{k-1}
	\right.
	\\
	&+
	\left.
	\bSig_k \bH_k^T \bV_k^{-1} \br_k)^T \right)
	\\
	\dot{L}_{d\bv_k}
	&=
	-
	\left(
	\dIAG{ \bH_k \bSig_k \bH_k^T}
	-\frac{1}{\alpha} \bv_k
	\right.
	\\
	&+
	\left.
	(\br_k - \bH_k\bSig_k\bH_k^T \bV_k^{-1}\br_k)^{\odot 2}
	\right)   \div\bv_k^{\odot 2} 
	\\
	\dot{L}_{d\Kab{\bX_k}{\bR}}
	&= \dot{L}_{d\bH_k} \Kab{\bR}{\bR}^{-1} -2\alpha \DIAG{\dot{L}_{d\bv_k}} \bH_k
	\\
	\dot{L}_{d\Kab{\bR}{\bR}}
	&= -\bH_k^T \left(\dot{L}_{d\bH_k} \Kab{\bR}{\bR}^{-1} -\alpha \DIAG{\dot{L}_{d\bv_k}} \bH_k \right)
	\\
	\dot{L}_{d\bs{k}_{\bX_k\bX_k}}
	&= \alpha \dot{L}_{d\bv_k}
	\\
	\dot{L}_{d\bLam_k}
	&=
	\bSig_k
	-\bSig_{k-1}
	+ 2 \bSig_{k-1} \bH_k^T \bS_k^{-1} \br_k \bet_{k-1}^T \bSig_{k-1}
	\\
	&+
	\bSig_k \bH_k^T \bV_k^{-1} \br_k \br_k^T \bV_k^{-1} \bH_k \bSig_k
	\\
	\dot{L}_{d\bet_k}
	&=
	-2\bSig_{k-1} \bH_k^T \bS_k^{-1} \br_k
	\\
	\dot{L}_{d\_d_n}
	&=
	2\sigma_n^2~\sumall{\dot{L}_{d\bv_k}}-2B\frac{1-\alpha}{\alpha}
	\end{align*}

	\textbf{\textit{Derivative Updates}}\\
	\textbf{Loop over $\theta_i \in \bthe$:}\vspace{-0.5cm}
	\begin{align*}
	\dot{\psi}_k &= \dot{\psi}_{k-1} -\frac{1}{2}\left( 
	\sumall{
		\dot{L}_{d\bet_k}
		\odot
		\dot{\bet}_{k-1}
	}
	\right.
	\\
	&+
	\left.
	\sumall{
		\dot{L}_{d\bLam_k}
		\odot
		\dot{\bLam}_{k-1}
	}
	+
	\sumall{
		\dot{L}_{d\Kab{\bR}{\bR}}
		\odot
		\dot{\bs{K}}_{\bR\bR}
	}
	\right.
	\\
	&+
	\left.
	\sumall{
		\dot{L}_{d\Kab{\bX_k}{\bR}}
		\odot
		\dot{\bs{K}}_{\bX_k\bR}
	}
	\right.
	\\
	&+
	\left.
	\sumall{
		\dot{L}_{d\bs{k}_{\bX_k\bX_k}}
		\odot
		\dot{\bs{k}}_{\bX_k\bX_k}
	}
	+
	\mathrm{1}_{[\bthe_k=\sigma_n]}\dot{L}_{d\_d_n}
	\right)\\
	\dot{\bH}_k &= \dot{\bs{K}}_{\bX_k\bR}\Kab{\bR}{\bR}^{-1}
	-\Kab{\bX_k}{\bR}\Kab{\bR}{\bR}^{-1}\dot{\bs{K}}_{\bR\bR} \Kab{\bR}{\bR}^{-1}; \\
	\dot{\bs{d}}_k&=\dIAG{\dot{\bs{K}}_{\bX_k\bX_k} - 
		\dot{\bs{K}}_{\bX_k\bR}\Kab{\bR}{\bR}^{-1}\Kab{\bR}{\bX_k}  
		\right.\\
		&
		\left.+\Kab{\bX_k}{\bR}\Kab{\bR}{\bR}^{-1} \dot{\bs{K}}_{\bR\bR} \Kab{\bR}{\bR}^{-1} \Kab{\bR}{\bX_k}
		\right.\\
		&
		\left.
		-\Kab{\bX_k}{\bR}\Kab{\bR}{\bR}^{-1}\dot{\bs{K}}_{\bR\bX_k}};
	\end{align*}\vspace{-0.9cm}
	\begin{align*}
	\dot{\bV}^{-1}_k &= - \mathrm{1}_{[\theta_i\neq\sigma_n]}
	\alpha\bV_k^{-1}  \DIAG{\dot{\bs{d}}_k}\bV_k^{-1}
	\\
	&-
	\mathrm{1}_{[\theta_i=\sigma_n]}
	2\sigma_n^2 \dot{\bV}^{-1}_k\dot{\bV}^{-1}_k
	;\\
	\dot{\bet}_k &= \dot{\bet}_{k-1} + \dot{\bH}_k^T \bV_k^{-1} \by_k
	+\bH_k^T \dot{\bV}^{-1}_k \by_k;\\
	\dot{\bLam}_k &= \dot{\bLam}_{k-1} +  \dot{\bH}_k^T \bV_k^{-1} \bH_k
	\\
	&+\bH_k^T \dot{\bV}^{-1}_k \bH_k
	+ \bH_k^T \bV_k^{-1} \dot{\bH}_k 
	\end{align*}

	For the noise $\sigma_n$, all kernel derivatives are zero, therefore the calculations simplify significantly.

	\section{Derivation of Recursive Collapsed Bound}
	\label{se:alt_view}
	We provide more details and a detailed derivation for the recursive collapsed lower bound \eqref{eq:recursive_lik} for the VFE model from Sect. \ref{sse:rec_grad_prop}. 
	Instead of lower bounding directly the batch log marginal likelihood as done by Titsias \cite{titsias2009variational} in the batch case, our approach relies on the recursive factorization of the
	joint log marginal likelihood
	$
	\log \pc{\by}{\bthe} 
	= \log \prod_{k=1}^K \pc{\by_k}{\by_{1:k-1},\bthe} 
	= \sum_{k=1}^K \log \pc{\by_k}{\by_{1:k-1},\bthe}.
	$
	The properties induced by the sparse augmented inducing point model yield
	$
	\log \pc{\by}{\bthe}=$ \\
	$\sum_{k=1}^K \log \int
	\pc{\by_k}{\bff_{k},\bthe} 
	\pc{\bff_k}{\bu,\bthe}  \pc{\bu}{\by_{1:k-1},\bthe} \diff{\bff_k} \diff \bu.
	$
	%
	We first introduce the variational distributions $ q_k(\bff_k,\bu)=\pc{\bff_k}{\bu,\bthe}q_k(\bu) \approx \pc{\bff_k,\bu}{\by_{1:k},\bthe}$, then by applying Jensen's inequality to each individual term in the true log marginal likelihood, we obtain the lower bound
	\begin{align*}
	\log \pc{\by}{\bthe}
	\geq \sum_{k=1}^K  \int
	\pc{\bff_k}{\bu,\bthe}q_k(\bu) \ldots
	\\
	\ldots\log
	\frac{\pc{\by_k}{\bff_{k},\bthe} \pc{\bff_k}{\bu,\bthe}  \pc{\bu}{\by_{1:k-1},\bthe} }
	{\pc{\bff_k}{\bu,\bthe}q_k(\bu)}
	\diff{\bff_k} \diff\bu.
	\end{align*}
	The quantity $\pc{\bu}{\by_{1:k-1},\bthe}$ is unknown, however, we can replace it with $ q_{k-1}(\bu)$ leading to 
	$\LL
	\left(q_1,\ldots,q_K,\bthe \right)$
	\begin{align}
	\begin{split}
	\label{eq:bound_rec}
	\sum_{k=1}^K  \int
	\pc{\bff_k}{\bu,\bthe}q_k(\bu)
	\log
	\frac{\pc{\by_f}{\bff_{k},\bthe}   q_{b-1}(\bu) }
	{q_{k}(\bu)}
	\diff{\bff_k} \diff\bu.
	\end{split}
	\end{align}
	Maximizing this lower bound 
	recursively with respect to the distributions $q_{k}(\bu) $  leads to a sequence of
	optimal variational distributions $q_k^*(\bu) $ for the inducing outputs 
	\begin{align}
	\label{eq:qi_star}
	\NN{\bu}{\bSig_k \left\{ \frac{1}
		{\sigma_n^2} \bH_k^T \by_k + \bSig_{k-1}^{-1}\bmu_{k-1}\right\},
		\bSig_k},
	\end{align}
	where $\bSig_k= \left( 
	\bSig_{k-1}^{-1} + \frac{1}
	{\sigma_n^2} \bH_k^T \bH_k
	\right)^{-1}$
	and $\bH_k=\Kab{\bX_k}{\bR}\Kab{\bR}{\bR}^{-1}$. 
	Plugging $q_k^*(\bu)$ into \eqref{eq:bound_rec} 
	yields 
	\begin{align}
	\begin{split}
	\label{eq:rec_collapsed}
	\LL_{REC}(\bthe)
	=
	\sum_{k=1}^{K} \left[
	\log \NN{\by_k}{\bH_k \bmu_{k-1},
		\ldots \right. \right. \\ \left. \left. \ldots
		\bH_k \bSig_{k-1} \bH_k^T + \sigma_n^2\II }
	-\frac{\TR{\bD_{\bX_k \bX_k}}}{2\sigma_n^2} \right]
	.
	\end{split}
	\end{align}
	The recursive bound $\LL_{REC}(\bthe)$ is equivalent to the batch collapsed bound $\LL_{VFE}(\bthe)$ in \eqref{eq:collapsed_bound} and it holds
	$\LL_{SVGP}(\bmu,\bSig,\bthe)\leq \LL_{REC}(\bthe)$ with equality when inserting the optimal variational posterior.
	The variational posterior update in \eqref{eq:qi_star} has the same form as the recursive updates in \eqref{eq:rec}. Similarly, the recursive collapsed lower bound in \eqref{eq:rec_collapsed}
	is equal to  \eqref{eq:recursive_lik}.

	We provide below a detailed derivation that follows closely 
	the proof in
	\cite{titsias2009variational} for the recursive collapsed bound \eqref{eq:rec_collapsed} as well as the sequence of optimal distributions \eqref{eq:qi_star}.
	We assume mini-batches of size $B$, that is, we have training data $\mathcal{D} = \left\{ \by_k, \bX_k \right\}_{k=1}^{K}$ and the corresponding latent function values$ \left\{ \bff_k\right\}_{k=1}^{K}$.
	We briefly recap the involved quantities and introduce abbreviations:
	\begin{align*}
	\pc{\by_k}{\bff_k,\bthe} &= \NN{\by_k}{\bff_k, \sigma_n^2\II};\\
	\pc{\bff_k}{\bu,\bthe} &= 
	\NN{\bff_k}{\bH_k \bu, \Kab{\bX_k}{\bX_k}- \bQ_{\bX_k \bX_k} };\\
	\pc{\bu}{\bthe} &= \NN{\bu}{\bs{0}, \Kab{\bR}{\bR} } = q_0(\bu);\\
	q_{k-1}(\bu) &= \NN{\bu}{\bmu_{k-1}, \bSig_{k-1}}
	\approx \pc{\bu}{\by_{1:k-1},\bthe}\\
	\bH_k & =\Kab{\bX_k}{\bR} \Kab{\bR}{\bR}^{-1};\\
	\bQ_{\bX_k \bX_k} &=  \Kab{\bX_k}{\bR} \Kab{\bR}{\bR}^{-1} \Kab{\bR}{\bX_k}.
	\end{align*}
	Starting from \eqref{eq:bound_rec}, we have the bound 
	\begin{align*}
	\sum_{k=1}^{K}  \int
	\pc{\bff_k}{\bu,\bthe} q_k(\bu)
	\log
	\frac{\pc{\by_k}{\bff_{k},\bthe}   q_{k-1}(\bu) }
	{q_k(\bu)}
	\diff{\bff_k} \diff\bu
	\end{align*}
	which can be  
	rearranged to
	\begin{align*}
	\sum_{k=1}^{K }
	\int q_k(\bu)  
	\left\{
	\log G(\bu,\by_k)
	+
	\log \frac{
		q_{k-1}(\bu)
	}{q_k(\bu)}
	\right\}   \diff{\bu},
	\end{align*}
	where 
	$\log G(\bu,\by_k)
	=\int \pc{\bff_k}{\bu,\bthe} \log 
	\pc{\by_k}{\bff_{k},\bthe}
	\diff{\bff_k}$.
	The integral involving $\bff_k$ is computed as $
	\log G(\bu,\by_k) = \int\pc{\bff_k}{\bu,\bthe} \log 
	\pc{\by_k}{\bff_{k},\bthe}
	\diff{\bff_k}
	$, 		which equals
	\begin{align*}
	&= \EEE{\bff_k\vert \bu}{
		-\frac{B}{2}\log (
		2\pi\sigma_n^2)
		%
		-\frac{1}{2}
		\left[\by_k-\bff_k \right]^T  \frac{1}{\sigma_n^2} \left[\by_k-\bff_k \right]
	}\\
	&= -\frac{B}{2}\log (
	2\pi\sigma_n^2 )
	-\frac{1}{2\sigma_n^2}\left(
	\by_k^T\by_k
	%
	+
	\EEE{\bff_k\vert \bu}{
		\bff_k^T\bff_k - 2\by_k^T\bff_k
	}
	\right).
	\end{align*}
	Using $\EE{\bx^T\bA\bx}=\TR{\bA\bSig}+\bmu^T\bA\bmu$ with $\p{\bx}= \NN{\bx}{\bmu,\bSig}$ yields
	\begin{align*}
	&\log G(\bu,\by_k) = -\frac{K}{2}\log (
	2\pi\sigma_n^2 )
	-\frac{1}{2\sigma_n^2}\left(
	\by_k^T\by_k 
	\right. \\ &\left.
	+
	\TR{\Kab{\bX_k}{\bX_k} -\bQ_{\bX_k \bX_k}} 
	%
	+ \bu^T \bH_k^T  \bH_k \bu  - 2 \by_k^T \bH_k \bu
	\right)\\
	&=  \log\left[ \NN{\by_k}{\bH_k \bu,\sigma_n^2 \II } \right] 
	%
	-\frac{1}{2\sigma_n^2}
	\TR{\Kab{\bX_k}{\bX_k} -\bQ_{\bX_k \bX_k}}
	.
	\end{align*}
	Substitute this expression back, the lower bound becomes
	\begin{align*}
	\begin{split}
	\sum_{k=1}^{K } \left[
	\int q_k(\bu) 
	\log \frac{
		\NN{\by_k}{\bH_k \bu,\sigma_n^2 \II }
		q_{k-1}(\bu)
	}{q_k(\bu)}
	\diff{\bu} 
	\right. \\ \left.
	-\frac{1}{2\sigma_n^2}
	\TR{\Kab{\bX_k}{\bX_k} -\bQ_{\bX_k \bX_k}}
	\right]. 
	\end{split}
	\end{align*}
	We can now maximize this bound with respect to $q_k(\bu)$. Here since we have not constrained $q_b$ to belong to any fixed family of distributions, we can compute the optimal bound by reversing the Jensen’s inequality leading to
	%
	\begin{align*}
	\LL_{REC}(\bthe) &=\sum_{k=1}^{K} \left[
	\log \int  
	\NN{\by_k}{\bH_k \bu,\sigma_n^2 \II }
	q_{k-1}(\bu)
	\diff{\bu}
	\right. \\  &\left.
	-\frac{1}{2\sigma_n^2}
	\TR{\Kab{\bX_k}{\bX_k} -\bQ_{\bX_k \bX_k}}  \right] \\
	&=
	\sum_{k=1}^{K } \left[
	\log   
	\NN{\by_k}{\bH_k \bmu_{k-1},\bH_k \bSig_{k-1} \bH_k^T + \sigma_n^2 \II }
	\right. \\  &\left.
	-\frac{1}{2\sigma_n^2}
	\TR{\Kab{\bX_k}{\bX_k} -\bQ_{\bX_k \bX_k}}  \right]
	\end{align*}
	where we used a linear Gaussian identity (see, e.g., \cite{Bishop2006} Ch. 2.3) in the last step . This is equal to Eq. \eqref{eq:recursive_lik}. 
	The optimal distribution $q_k^*$ that gives rise to this bound is proportional to 
	$\NN{\by_k}{\bH_k \bu,\sigma_n^2 \II }
	q_{k-1}(\bu) 
	=
	\NN{\by_k}{\bH_k \bu,\sigma_n^2 \II} \NN{\bu}{\bmu_{k-1},\bSig_{k-1}}
	$ 
	and can be analytically computed 
	leading to
	\[
	q_k^*(\bu) = \NN{\bu}{\tfrac{1}
		{\sigma_n^2}\bSig_k \bH_k^T \by_k + \bSig_{k-1}^{-1}\bmu_{k-1},\bSig_k}
	\]
	where $\bSig_k= \left( 
	\bSig_{k-1}^{-1} + \frac{1}
	{\sigma_n^2} \bH_k^T \bH_k
	\right)^{-1}$.
	This matches the result in Eq. \eqref{eq:rec} and \eqref{eq:KF}
	and completes the proof.

	\section{Proof of proposition \ref{prop:gradients}}
	We showed (App.~\ref{se:alt_view}) how the batch approximate log-likelihood $\LL_{PEP}$ in \eqref{eq:PEP_log} can be recursively computed. In Sect. \ref{sse:rec_grad_prop}, we showed an equivalent way to compute it, that is, we have the cumulative bound $\psi^{(K)}(\bthe)=\sum_k^K\psi_k(\bthe)$ and it satisfies $\LL_{PEP}(\bthe)=\sum_k^K\psi_k(\bthe)$. By induction and linearity of derivatives, we therefore get $\parD{\psi^{(K)}(\bthe)}{\theta}=\parD{\psi_k(\bthe)}{\theta} + \parD{\psi^{(K-1)}(\bthe)}{\theta}$ and $\parD{\LL_{PEP}}{\theta}=\parD{\psi^{(K)}}{\theta}$.
	
\end{document}